\documentclass[journal, twoside]{IEEEtran}
\usepackage{times}
\usepackage[pdftex]{graphicx}

\usepackage{amsmath,amssymb,amsopn,amstext,amsfonts, amsthm}
\usepackage{cancel}
\usepackage[space]{cite}
\usepackage{pdfsync}
\usepackage{balance}
\usepackage{color}
\usepackage{mathtools}
\usepackage[ruled,norelsize, linesnumbered]{algorithm2e}
\usepackage{bm}
\usepackage{diagbox}
\usepackage{float}
\usepackage[outdir=./]{epstopdf}
\usepackage{url}
\usepackage{multirow}
\usepackage{makecell}
\usepackage{adjustbox}
\usepackage{tabularx, booktabs}

\usepackage{enumitem}
\usepackage[font=small,skip=2pt]{caption}
\usepackage[linkcolor=black,citecolor=black,urlcolor=black,colorlinks=true]{hyperref}
\usepackage{subcaption}
\usepackage{bookmark}
\usepackage{pifont}

\newcommand{\cmark}{\ding{51}}
\newcommand{\xmark}{\ding{55}}

\theoremstyle{remark}

\newcommand{\func}{\mathtt}

\DeclareMathOperator*{\argmax}{arg\,max}

\SetKw{Continue}{continue}
\makeatletter
\newcommand{\removelatexerror}{\let\@latex@error\@gobble}
\makeatother

\graphicspath{{./}}
\DeclareGraphicsExtensions{.pdf,.png,.jpg,.eps}

\begin{document}

\title{EPSILON: An Efficient Planning System \\ for Automated Vehicles in Highly \\ Interactive Environments}
\author{Wenchao Ding, Lu Zhang, Jing Chen, and Shaojie Shen
	\thanks{Manuscript received: December 4, 2020; Revised May 1, 2021; Accepted July 12, 2021. This paper was recommended for publication by Editor Fran\c{c}ois Chaumette upon evaluation of the Associate Editor and Reviewers' comments. This work was supported in part by the Hong Kong Ph.D. Fellowship Scheme, in part by the HKUST-DJI Joint Innovation Laboratory, and in part by the HKUST Institutional Fund. (\textit{Wenchao Ding and Lu Zhang contributed equally to this work. Corresponding author: Lu Zhang}.)}
	\thanks{Wenchao Ding is with the Huawei Technology Co., Ltd., Shanghai, China, and the work was done while he was at the Hong Kong University of Science and Technology, Hong Kong, China (email: wdingae@ust.hk).	Lu Zhang and Shaojie Shen are with the Department of Electronic and Computer Engineering, Hong Kong University of Science and Technology, Hong Kong, China (email: lzhangbz@ust.hk; \mbox{eeshaojie@ust.hk}). Jing Chen is with the DJI Technology Company, Ltd., Shenzhen 510810, China (email:jing.chen@dji.com).}
	\thanks{This paper has supplementary downloadable multimedia material available at \url{http://ieeexplore.ieee.org}.}
	\thanks{Color versions of one or more of the figures in this paper are available online at \url{http://ieeexplore.ieee.org}.}
	\thanks{Digital Object Identifier (DOI): TBD}
}

\maketitle

\begin{abstract}
	In this paper, we present an \underline{E}fficient \underline{P}lanning \underline{S}ystem for automated vehicles \underline{I}n high\underline{L}y interactive envir\underline{ON}ments (EPSILON). EPSILON is an efficient interaction-aware planning system for automated driving, and is extensively validated in both simulation and real-world dense city traffic. It follows a hierarchical structure with an interactive behavior planning layer and an optimization-based motion planning layer. 
	The behavior planning is formulated from a partially observable Markov decision process (POMDP), but is much more efficient than naively applying a POMDP to the decision-making problem. The key to efficiency is guided branching in both the action space and observation space, which decomposes the original problem into a limited number of closed-loop policy evaluations.
	Moreover, we introduce a new driver model with a safety mechanism to overcome the risk induced by the potential imperfectness of prior knowledge. 
	For motion planning, we employ a spatio-temporal semantic corridor (SSC) to model the constraints posed by complex driving environments in a unified way. Based on the SSC, a safe and smooth trajectory is optimized, complying with the decision provided by the behavior planner. 
	We validate our planning system in both simulations and real-world dense traffic, and the experimental results show that our EPSILON achieves human-like driving behaviors in highly interactive traffic flow smoothly and safely without being over-conservative compared to the existing planning methods.
\end{abstract}


\section{Introduction}\label{sec:introduction}
Autonomous driving is an emerging topic, both in industry and in the academic community. Planning, as one of the core components of autonomous driving systems, largely determines the intelligence and user experience of the system. Despite there being many industrial demos illustrating promising autonomy, few methodological details are provided to show how pain points in planning are systematically dealt with. On the other hand, the academic community has presented various planning systems which cover several pain points, such as uncertainty and interaction modeling~\cite{ding2019safe,hubmann2017decision,xu2012optimization,liu2017speed,zhu2015convex,Gu2013Focused,wei2014behavioral,ajanovic2018search,zhan2017spatially}. However, most of these methods are only validated through simulation or well-annotated datasets, leaving a question as to whether they can work on a real autonomous vehicle. Planning on a real vehicle with closed-loop execution is far more challenging. Onboard planning requires dealing with an imperfect world and interacting with other traffic participants naturally. The imperfectness and uncertainty come from a wide range of aspects, such as occlusion in sensing, detection and tracking noises, and unavoidable stochastic behaviors of other traffic participants, which are hard to reproduce in simulation. In this paper, we aim at building a robust and socially-compliant planning system which can handle the imperfect real world.

In this paper, we present an \underline{E}fficient \underline{P}lanning \underline{S}ystem for autonomous vehicles \underline{I}n high\underline{L}y interactive envir\underline{ON}ments (EPSILON). We systematically investigate several pain points of planning for autonomous driving, such as the modeling of uncertainty and multi-agent interaction, and attempt to achieve one small but concrete step towards automated driving in the real-world. We validate the performance of EPSILON by conducting long-term closed-loop autonomous driving in complex driving environments. EPSILON follows a hierarchical structure which consists of a behavior planning layer and a motion planning layer, as in many previous methods~\cite{zhan2017spatially,wei2014behavioral,ziegler2014making}. In EPSILON, the role of behavior planning is to generate a preliminary decision which is represented by a sequence of states covering the planning horizon, while the role of motion planning is to wrap the sequence of states to a safe and smooth trajectory for closed-loop execution. 

There is an extensive literature on behavior planning for automated vehicles. Previous works~\cite{montemerlo2008junior,urmson2008autonomous,ziegler2014making,hubmann2016generic,ajanovic2018search,wei2014behavioral,zhan2017spatially,fan2018baidu} focused on a pure geometry perspective, and adopted various rule-based or search-based techniques, which typically assume a behavior/trajectory prediction module is equipped on the system. The prediction is conducted independent of planning, and when the prediction is provided, the planner plans a trajectory with a sufficiently large safety margin.
However, it is impossible to yield a perfect prediction due to the stochastic nature of other traffic participants and onboard perception noise.
Moreover, with such independent prediction, the mutual influence of the ego vehicle's and other vehicles' future motion cannot be modeled. In cooperative and interactive scenarios, such as gap merging, where this mutual influence is essential, coupled prediction and planning is more promising~\cite{naumann2020irl}.
Partially observable Markov decision process (POMDP)~\cite{kaelbling1998planning} provides a mathematically rigorous form of modeling uncertainties and multi-agent interactions, while suffering from prohibitively high computational complexity. Several attempts~\cite{kurniawati2016abt,ye2017despot} have been made to accelerate the problem solving but are still not efficient enough for driving in complex environments~\cite{hubmann2018belief,bai2015intention}. 
In this paper, we propose a guided branching technique to focus the exploration using domain knowledge, which is much more efficient and can work in highly dynamic city traffic.

Motion planning aims at generating a safe and smooth trajectory which faithfully follows the decision provided by the behavior layer. Our motion planning layer is adapted from our previous work~\cite{ding2019safe}, which follows an optimization-based scheme. All the constraints posed by complex semantics are encoded in a unified way using a spatio-temporal semantic corridor. The piecewise B\'{e}zier curve is adopted as trajectory parameterization for its convex hull and hodograph properties which can enforce safety and dynamical feasibility for the entire trajectory. Benefiting from our optimization formulation, our motion planner can generate a safe and smooth trajectory as well as fit the preliminary behavior plan closely, significantly enhances the consistency of EPSILON.

The behavior planning layer of EPSILON was originally presented in our previous research~\cite{ding2020eudm}. In~\cite{ding2020eudm}, although the interaction among traffic participants is captured using multi-agent forward simulation, it generally assumes other participants are \textit{rational}. The rationality is reflected in the predefined multi-agent integration model. However, in real-world city driving, we find traffic participants are often \textit{noisily rational}, especially in cities where the driving style is always aggressive. 
In this paper, we extend our previous work by integrating a more flexible interactive forward simulation model with a safety mechanism, which better guarantees safety, even when encountering over-aggressive traffic participants. The motion planning module of EPSILON was originally proposed in~\cite{ding2019safe}. We advance the motion planner to incorporate multi-agent interaction and unify behavior planning and motion planning. Moreover, we bring the newly designed system, EPSILON, into real-world dense traffic, while~\cite{ding2019safe} and~\cite{ding2020eudm} are limited to only the simulation environment. We summarize our contributions as follows:
\begin{enumerate}
	\item We present an efficient planning system by extending and tightly integrating our previous behavior planner~\cite{ding2020eudm} and motion planner~\cite{ding2019safe}. The proposed new system deals with several pain points of automated driving systematically, including efficient interaction and uncertainty handling, improved consistency, and real-time implementation.
	\item We enhance the robustness of the behavior planning layer by introducing a new forward simulation model, which better guarantees safety, even when encountering over-aggressive or uncooperative traffic participants.
	\item Moving beyond validation using only simulation and datasets, we extensively validate our system on a real automated vehicle in dense city traffic without an HD-map and purely relying on onboard sensor suites.
	\item We release the complete decision-making and motion planning systems as open-source packages\footnote{https://github.com/HKUST-Aerial-Robotics/EPSILON} to the research community.
\end{enumerate}

The remainder of this paper is organized as follows. The relevant literature is discussed in Section~\ref{sec:related_work}. An overview of EPSILON is provided in Section~\ref{sec:system_overview}. The problem is formulated in Section~\ref{sec:prob}. The behavior planning module is elaborated in Section~\ref{sec:bp} while the motion planning method is presented in Section~\ref{sec:ssc_traj}. Implementation details are given in Section~\ref{sec:implementation}. Systematic comparisons and real-world experiments are illustrated in Section~\ref{sec:experimental_results}. Finally, Section~\ref{sec:conclusion} concludes this paper.

\section{Related Work}\label{sec:related_work}
\noindent \textbf{Behavior planning for automated vehicles:} A significant number of works on behavior planning\footnote{In this paper, the terms ``behavior planning'' and ``decision-making'' are used interchangeably.} for automated vehicles have been published in recent years~\cite{schwarting2018planning}, and many of them consider the problem from a geometric perspective, namely, finding preliminary geometric collision-free paths/trajectories for the controlled vehicle. State-machine-based~\cite{montemerlo2008junior,urmson2008autonomous,ziegler2014making} behavior planning, which employs handcrafted rules for different driving conditions, was popular at the early stage. In this approach, based on the output state and action, reference paths can be extracted and fed to the motion planning layer. However, due to the complexity of real-world driving, continuous engineering and tuning efforts are required to maintain the state machine. For this reason, later methods turned to a more generic formulation for behavior planning, such as search-based methods~\cite{ajanovic2018search,hubmann2016generic,wei2014behavioral,zhan2017spatially,fan2018baidu}. For example, Hubmann et al.~\cite{hubmann2016generic} used A$^\ast$ graph search on a state lattice with static and dynamic events encoded as a cost map.
The methods which adopt a purely geometric reasoning typically assume a deterministic trajectory prediction of other traffic participants is provided for the purpose of collision-checking. However, in real-world driving, prediction should be modeled from a probabilistic perspective due to the multi-modal and uncertain nature of the prediction problem. Moreover, different future actions of the controlled vehicle may result in different future situations, which is also not modeled under purely geometric reasoning. In this paper, we tackle the behavior planning problem from an interaction-aware perspective, which fundamentally addresses the above issues.

There is extensive literature on behavior planning from an interactive multi-agent perspective, and many of the methods are formulated as POMDPs~\cite{kaelbling1998planning}. The POMDP is mathematically rigorous and outlines a principled way to solve the problem of planning under uncertainty. Due to the \textit{curse of dimensionality}, the POMDP quickly becomes computationally intractable when the problem size scales. To overcome this issue, many online POMDP solvers have been proposed to accelerate the problem solving, such as those in~\cite{kurniawati2016abt,silver2010pomcp,ye2017despot,cai2018hyp}. Leveraging the latest advances in POMDP solvers, several works~\cite{hubmann2018automated,bai2015intention,hubmann2018belief,hubmann2017decision} have applied POMDPs to behavior planning for automated vehicles. However, most conduct validation only in simulation with particular scenarios, such as merging or intersections, except~\cite{bai2015intention}, which conducts onboard experiments in a crowd. Nevertheless,~\cite{bai2015intention} only deals with a tailored 1-D problem, namely, the speed optimization problem under a given path but yielding a limited efficiency (around 3 to 5 Hz), which may be inadequate for solving a full decision-making problem in highly dynamic driving environments. One solution to handle the computational issue is to simplify the original POMDP using domain knowledge. One representative method, multiple policy decision making (MPDM), was proposed in~\cite{cunningham2015mpdm,galceran2017mpdmar}. It firstly designs a set of semantic-level policies such as lane-nominal and yield, then conducts closed-loop forward simulation to evaluate the policies. MPDM converts the original problem into selecting the best policy with highest reward in a limited number of policies which is computationally efficient and easy to deploy. However, MPDM over-simplifies the action for the ego vehicle: each candidate policy contains only one semantic-level action with a rather long duration, making MPDM a single-stage MDP, which restricts the manoeuvrability and the level of ``intelligence'' of the decision. In this paper, we retain the multiple layer structure of the POMDP and utilize domain knowledge as well as guided branching to achieve higher efficiency while retaining flexibility. Moreover, our method is validated against real-world environments with rich interactions and potentially stochastic traffic participants.

\noindent \textbf{Motion planning for automated vehicles:} Motion planning for automated vehicles is relatively mature compared to behavior planning~\cite{gonzalez2015review, paden2016survey}. Typical routines for motion planning include search-based methods based on the state lattice~\cite{ziegler2009spatiotemporal,martin2010On,mcnaughton2011motion} or motion primitives~\cite{werling2012mp}, optimization-based methods~\cite{wolf2008artificial,xu2012optimization,liu2017speed,zhu2015convex,ziegler2014local}, or the combination of these two~\cite{Gu2013Focused,gu2015tunable}. Besides, works focus on higher safety guarantee that utilizes the reachability analysis to generate a trajectory under the constraints of reachable set that considers the uncertainties of the vehicle models~\cite{althoff2014online, vaskov2019guaranteed}, limited sensing~\cite{bajcsy2019efficient}, and interactions~\cite{leung2020infusing}.
Our previous method~\cite{ding2019safe} belongs to the category of optimization-based methods. The key feature is that it proposes a unified input representation, namely, the spatio-temporal semantic corridor structure, to wrap complex constraints posed by various semantic elements. This input representation saves considerable effort in engineering and tuning the constraints. However, like most motion planning methods, our previous approach still requires a trajectory prediction module and does not capture the multi-agent interaction, which poses inconsistency when being used with interaction-aware behavior planning methods. In this paper, since future anticipation is coupled inside our behavior planning, we can directly feed the decision together with the conditional anticipation of other traffic participants to the motion planning layer, which makes the motion planner in~\cite{ding2019safe} amenable to modeling interaction.

\section{System Overview}\label{sec:system_overview}
\begin{figure}[t]
	\centering
	\includegraphics[width=0.48\textwidth]{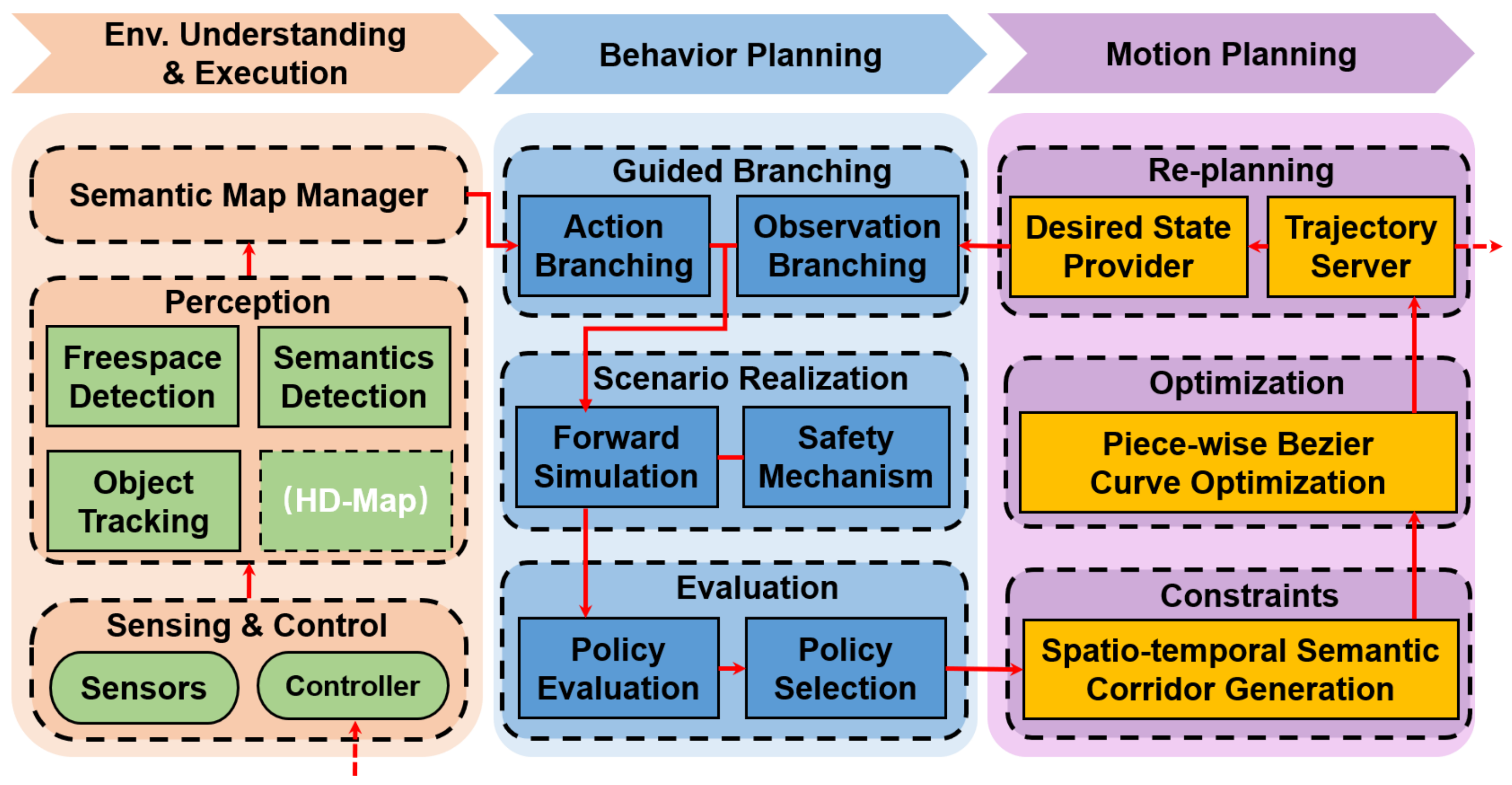}
	\caption{A diagram of EPSILON together with environmental understanding and execution modules. HD-Map is optional for EPSILON.\label{fig:system_overview}}
	\vspace{-1.0cm}
\end{figure}

The structure of the proposed planning system is shown in Fig.~\ref{fig:system_overview}. The planning system is built on top of the environment understanding and closed-loop execution modules. There are several modules in the perception pipeline: freespace detection~\cite{zhao2020fp}, semantics detection (e.g., lane detection, traffic light detection, etc.), object detection and tracking. Note that HD-map is \textit{optional} for our planning system. For onboard experiments on a real vehicle, we do not use HD-Map and purely rely on online lane detections. The output of perception is synchronized and fed to a semantic map manager module which is responsible for organizing the data structures and providing querying interfaces for planning modules. The semantic map is updated at $20$ Hz and includes an occupancy grid for static obstacles, multiple tracklets for dynamic obstacles, and road structures for lanes and traffic semantics.

As mentioned previously, EPSILON follows a two-layer hierarchical structure, comprising a behavior planning layer and a motion planning layer. Note that there is no additional trajectory prediction module like in~\cite{ding2019safe} since the trajectory predictor is coupled inside the behavior planner. At a high level, our behavior planner consists of three processes, namely, guided branching, scenario realization and evaluation. Essentially, guided branching is responsible for expanding action sequences according to predefined policies for the controlled vehicle and reasoning about the possible intentions of other traffic participants. By incorporating a particular ego action sequence with a particular intention combination of other traffic participants, we form a \textit{scenario}. During scenario realization, we use closed-loop multi-agent forward simulation to realize the scenario step-by-step. Different from our previous work~\cite{ding2020eudm}, we extend the forward simulation model into a more flexible driving model to achieve safe and human-like driving behavior, even in complex real-world traffic with quantities of noisily rational participants.

The motion planning layer basically follows our previous work~\cite{ding2019safe}. Firstly, static, dynamic obstacles and constraints posed by environment semantics are modeled using a spatio-temporal semantic corridor around the initial guess provided by the behavioral layer. Then a piece-wise B\`{e}zier curve is optimized with respect to the corridor. Using its convex hull property and hodograph property, the \textit{entire} trajectory can be guaranteed to be safe and dynamically feasible. The generated trajectory is fed to a trajectory server for re-planning scheduling, and the desired state for re-planning can be queried from the trajectory server. The trajectory server is also responsible for sending out control commands to a vehicle controller to close the execution loop.

The design frequency for both layers is 20 Hz. In practice, the two layers can be assembled as a pipeline, which will increase the throughput of the whole planning system.

\begin{figure}[t]
	\centering
	\includegraphics[width=0.35\textwidth]{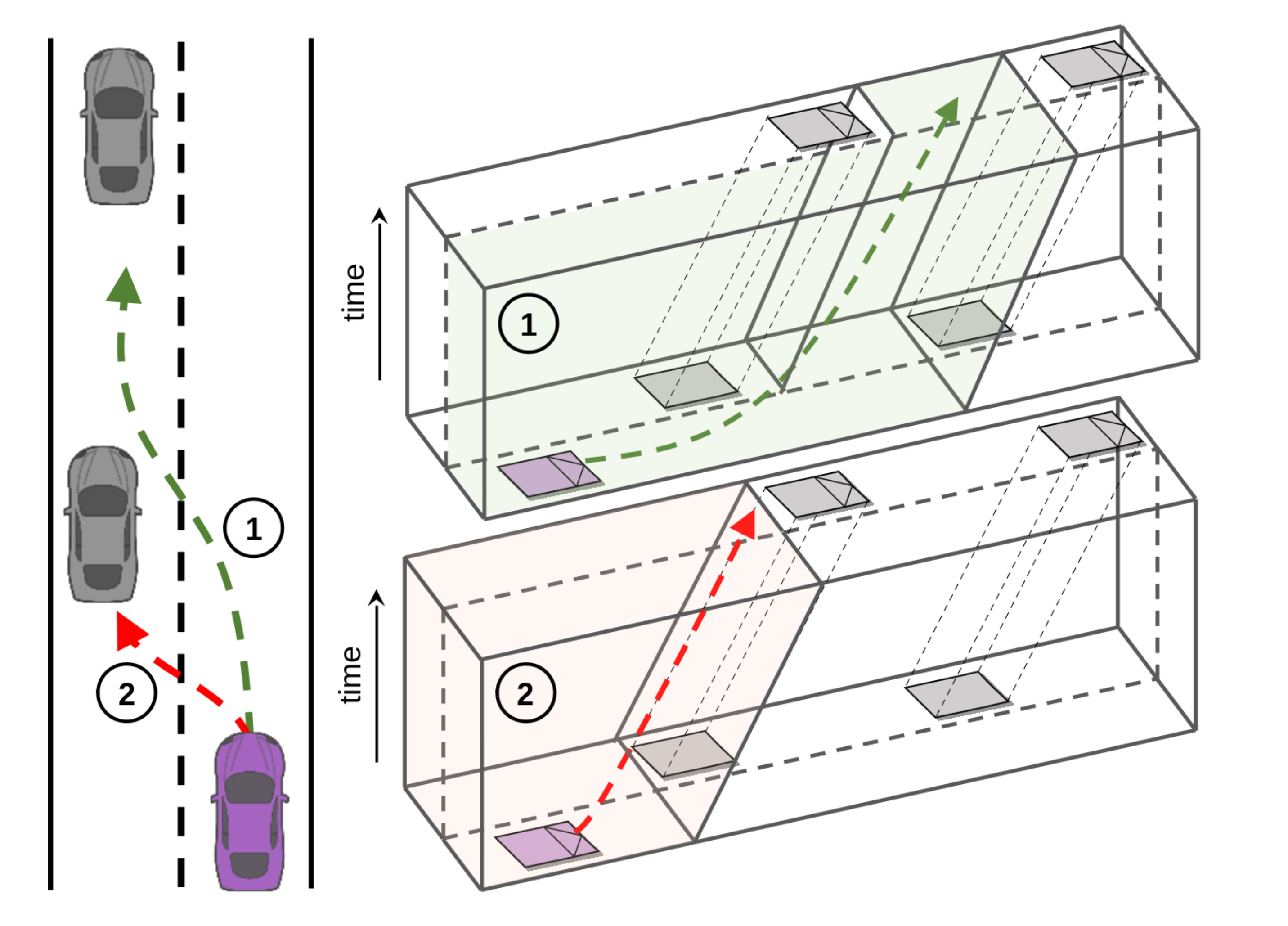}
	\caption{Illustration of the relationship between behavior planning and motion planning defined in this paper. The controlled vehicle (marked in \textit{purple}) can accelerate and insert between two nearby vehicles (maneuver \textcircled{\raisebox{-0.9pt}{1}}), or yield and follow the nearest vehicle (maneuver \textcircled{\raisebox{-0.9pt}{2}}). Visualized in the spatio-temporal domain, the two maneuvers belong to two different homotopy classes. The behavior planning targets at discovering diverse tactical maneuvers, while the motion planning is for generating a local smooth and safe trajectory within the corresponding spatio-temporal space.\label{fig:bp_mp_relationship}}
\end{figure}

\section{Problem Formulation}\label{sec:prob}

We firstly clarify the problem setup of the decision making and motion planning process of the ego agent. Let $\mathcal{E}_t$ denote the ego-centric local environment at time $t$, including road structures, traffic signals and the occupancy grid for static obstacles. Define $x^i_t$ as the state of the vehicle $i\in V$ at time $t$, and without loss of generality, $i=0$ is reserved for the ego vehicle. As a notational convenience, subscript absence denotes \textit{all} time steps, and superscript absence denotes all agents. For example, $x_t$ denotes the states of all vehicles at time $t$, while $x^i$ denotes the states at all time instances for vehicle $i$. For the planning cycle at time $t$, the ego agent receives the observation $z_t$, and uses it to estimate the real state $x_t$ for planning. We define the input of the behavior planning layer as $\langle z_t, \mathcal{E}_t \rangle$, and the output is a \textit{decision} $\mathcal{D}_t$, which is parameterized by a sequence of discrete states $\mathcal{D}_t \coloneqq [x_{t+1}, x_{t+2},\ldots, x_{t+H}]$ for all agents, where $H$ denotes the planning horizon. The input to the motion planning layer is $\langle \mathcal{D}_t, \mathcal{E}_t \rangle$, while the output is a smooth and safe trajectory (typically parameterized by splines) for controlling the ego vehicle. The reason that we decouple the behavior planning and motion planning layer is for higher efficiency. After decoupling, the behavior planner only needs to reason about the future scenario at a relatively coarse resolution, while the motion planning layer works in the local solution space given the decision $\mathcal{D}_t$.

As mentioned in Section~\ref{sec:introduction}, in the real world, the planning system always suffers from unknown interactive patterns among traffic participants. To address this issue, we model the uncertainty-aware behavior planning problem in the form of the POMDP, after which we formulate it into a multi-agent setting. A POMDP model can be defined as a tuple $\langle\mathcal{X}, \mathcal{A},  \mathcal{Z}, T, O, R \rangle$, where $\mathcal{X}$, $\mathcal{A}$ and $\mathcal{Z}$ are the state space, action space and observation space, respectively. The function $T(x_{t-1}, a_t, x_t) = p(x_t|x_{t-1}, a_t)$ is the probabilistic state transition model, while $O(x_t, z_t)=p(z_t|x_t)$ is the observation model. These two functions reflect the stochastic property of the motion model and uncertain sensing. $R(x_{t-1},a_t)$ is a real-valued reward function $R:\mathcal{X}\times\mathcal{A}\rightarrow\mathbb{R}$ for the agent taking action $a_t\in\mathcal{A}$ in state $x_{t-1}\in\mathcal{X}$. Since some states in real-world applications cannot be directly observed (e.g., hidden intentions or noisy measurements), the POMDP maintains the \textit{belief} $b\in\mathcal{B}$, which is the probability distribution over $\mathcal{X}$. The belief can be updated after the agent leaves an initial belief $b_{t-1}$, and then takes an action $a_t$ and receives an observation $z_t$. The resulting belief state can be inferred using Bayes' rule:
\begin{align}
    b_t(x_{t}) &= p(x_{t}|z_t, a_t, b_{t-1})\nonumber\\
               &= \eta O(x_{t},z_t)\int_{x_{t-1}\in\mathcal{X}}T(x_{t-1},a_{t},x_t)b_{t-1}(x_{t-1})dx_{t-1},
    \label{eq:belief_update}
\end{align}
where $\eta$ is a normalizing factor. 
The POMDP is to find an optimal \textit{policy} $\pi^*$ which maps the belief state to an action $\pi:\mathcal{B}\rightarrow\mathcal{A}$, maximizing the expected total discounted reward over the planning horizon:
\begin{equation*}
	\pi^{\ast} \coloneqq \underset{\pi}{\argmax}\mathbb{E} \left[\sum_{t=t_0}^{t_H}\gamma^{t-t_0}R(x_{t},\pi(b_{t}))\big\vert b_{t_0} \right],
\end{equation*}
where $t_0$ is the current planning time and $\gamma\in[0,1]$ is a discount factor. For online POMDPs, starting from an initial belief $b_{t_0}$ and expanding in action space $\mathcal{A}$ as well as observation space $\mathcal{Z}$ until a certain planning horizon $t_H$, a belief tree is built node by node. Then, an optimal policy is found by applying the Bellman equation on each internal node:
\begin{align*}
    V^*(b)=\max_{a\in\mathcal{A}}&Q^*(b,a)\\
        =\max_{a\in\mathcal{A}}&\Big\{ \int_{x\in\mathcal{X}}b(x)R(x,a)dx\\
     & + \gamma\int_{z\in\mathcal{Z}}p(z|b,a)V^*\big(\tau(b,a,z)\big)dz\Big\},
\end{align*}
where $V^*(b)=\int_{s\in\mathcal{S}}V^*(s)b(s)ds$ is the optimal utility function for the belief state, and $Q^*(b,a)$ describes the optimal value of the belief-action pair. For more details about solving POMDPs, we refer interested readers to~\cite{kaelbling1998planning, silver2010pomcp}.

Different from previous works~\cite{bai2015intention,hubmann2018automated} that only use the single optimal action on the initial belief node as the final output, here, we extract a complete trace $\mathcal{S}_t=[b^*_{t}, a^*_{t+1}, z^*_{t+1}, b^*_{t+1},...,b^*_{t+H}]$ on the belief tree by applying  
\begin{align*}
    a^*_{t} &= \argmax_{a_t\in\mathcal{A}}Q^*(b_{t-1},a_t) \\
    z^*_{t} &= \argmax_{z_t\in\mathcal{Z}}p(z_t|b_{t-1},a^*_t)
\end{align*}
on action branches and observation branches recurrently. The resulting trace contains the optimal action $a^*_t$ on each belief node $b^*_{t-1}$, and the most likely received observation $z^*_t$ given $a^*_t$. The final decision $\mathcal{D}_t$ can be generated by simply applying $x_t=\argmax_{x}b^*_t(x)$ for each step in $\mathcal{S}_t$. Compared to a single optimal action, $\mathcal{D}_t$ contains much more holistic information about the environment and future forecasting, which are essential for the following motion planner.

Considering the driving scenario at time $t$ with $N$ agents, the full state can be written as $x_t=\{x^0_t, x^1_t, ..., x^N_t\}\in\mathcal{X}$, wherein $x^i_t\in\mathcal{X}^i$ is the $i$-th vehicle's state containing its position, velocity, acceleration, heading, and steering angle. Note that for surrounding vehicles, hidden states such as driving \textit{intention} or \textit{aggressiveness}, which can not be directly observed, are also included in $x^i_t, \forall i\neq0$. In the driving scenario, the only thing we can control is the input signal (e.g., throttle/brake and steering) of the ego vehicle; in other words, we cannot directly determine the surrounding agents' actions. Therefore, we point out that the action of the original problem can be written as $a_t=a^0_t$, similarly, the action space $\mathcal{A}$ is equivalent to $\mathcal{A}^0$. The complete state transition model $T$ can be represented as a joint distribution of the full state $p(x_{t}|x_{t-1},a_t)=p(x^{0}_{t},...,x^{N}_{t}|x^{0}_{t-1},...,x^{N}_{t-1},a^{0}_t)$ which is non-trivial to directly model. However, since most of the agents in real traffic, such as vehicle and bicycle, follow common traffic rules and physical principles (e.g., kinematics or dynamics), we can simplify the formulation by making reasonable assumptions for the surrounding agents and converting it into a multi-agent interactive model. We assign a probabilistic transition model with an action $a^{i}_t\in\mathcal{A}^i, \forall i\neq0$ for each surrounding agent and assume that the instantaneous transitions of each vehicle are independent, resulting in
\begin{align}
    &p(x_{t}|x_{t-1},a_t)\nonumber \\
    \approx & 
    \underbrace{p(x^0_{t}|x^0_{t-1},a^0_t)}_{\text{ego transition}}
    \prod^{N}_{i=1}\int\limits_{\mathcal{A}^i}
    \underbrace{p(x^i_{t}|x^i_{t-1},a^i_{t})}_{\text{$i$-th agent's transition}}
    \underbrace{p(a^i_{t}|x_{t-1})}_{\text{driver model}}da^i_t, \label{eq:transition_decomp}
\end{align}
which separates the controlled vehicle from the surrounding agents. In the formulation, $p(x^i_{t}|x^i_{t-1},a^i_{t})$ is the \textit{assumed} transition model of the other agents that reflects the agent's low-level kinematics and $p(a^i_{t}|x_{t-1})$ is the \textit{assumed} driver model representing the high-level decision process of the other agents that provides appropriate control signals according to the driving context. In practice, the driver model can be user-defined or learned by data-driven approaches, and can be controlled by some latent states, such as intentions or aggressiveness, achieving diverse \textit{maneuvers} or \textit{driving styles}. Note that the ego action $a^0_t$ is generated by the predefined ego policy. The ego observation $z^0_t=z_t$ contains the estimated position and velocities of other vehicles generated by the perception module. For the surrounding agents, we use the agents' poses as the origin and transform the ego observation $z^0_t$ into their coordinate frames as observation $z^i_t$. Since $z^i_t$ is fully determined by the ego observation, we can get $p(z_t|x_t)=\prod^{N}_{i=0}p(z^i_t|x^i_t)$ assuming the observation processes are independent. Therefore, Eq.~\ref{eq:belief_update} can be factorized as 

\begin{small}
\begin{align}
    &b_t(x_{t})
    = \eta\cdot 
    \overbrace{p(z^0_t|x^i_{t})\int\limits_{\mathcal{X}^0}p(x^0_{t}|x^0_{t-1},a^0_{t})b^0_{t-1}(x^0_{t-1})dx^{0}_{t-1}}^{\text{belief update for ego agent}}\nonumber \\
    &\prod^{N}_{i=1}
    \overbrace{p(z^i_t|x^i_{t})\iint\limits_{\mathcal{X}^i\mathcal{A}^i}p(x^i_{t}|x^i_{t-1},a^i_{t})p(a^i_{t}|x_{t-1})b^i_{t-1}(x^i_{t-1})da^{i}_{t}dx^{i}_{t-1}}^{\text{belief update for other agents}}. \label{eq:belief_update_new}
\end{align}
\end{small}

We find that although the state transitions of each agent in each step are independent, the assumed driver models $p(a^i_t|x_{t-1})$ and observation models $p(z^i_t|x^i_{t})$ leverage all agents' states and observations, making the belief update an interactive process. Note that the hidden states are estimated step-by-step during the belief update using observations, which is exactly the role of behavior prediction. As a consequence, the POMDP implicitly contains the prediction, indicating that planning and prediction are naturally coupled. It is also notable that methods which decouple prediction and planning are essentially simplifications of the original POMDP formulation.

By applying belief update recurrently, we can build the belief tree starting from the initial belief node and extract the final decision $\mathcal{D}_t$. However, the scale of the belief tree grows exponentially w.r.t. the tree depth, which is computationally intractable for real-time applications. To overcome this issue, in this paper, we apply domain knowledge into the formulation to further simplify the problem of achieving real-time decision-making in a fast-changing driving environment while preserving the ability to handle interactions and uncertainties. Given the output of behavior planner $\mathcal{D}_t$, the motion planning layer aims to generate a safe and smooth trajectory to realize the high-level decision in a fine-grained fashion. Note that since our behavior layer is formulated in a multi-agent setting, it naturally reasons about the future states for \textit{all} the agents. Therefore, the role of the motion planning boils down to a local trajectory optimization problem, as shown in Fig.~\ref{fig:bp_mp_relationship}.

\section{Efficient Behavior Planning}\label{sec:bp}
\subsection{Motivating Example}\label{sec:bp_motivating}
\begin{figure}[tb]
	\centering
	\includegraphics[width=0.35\textwidth]{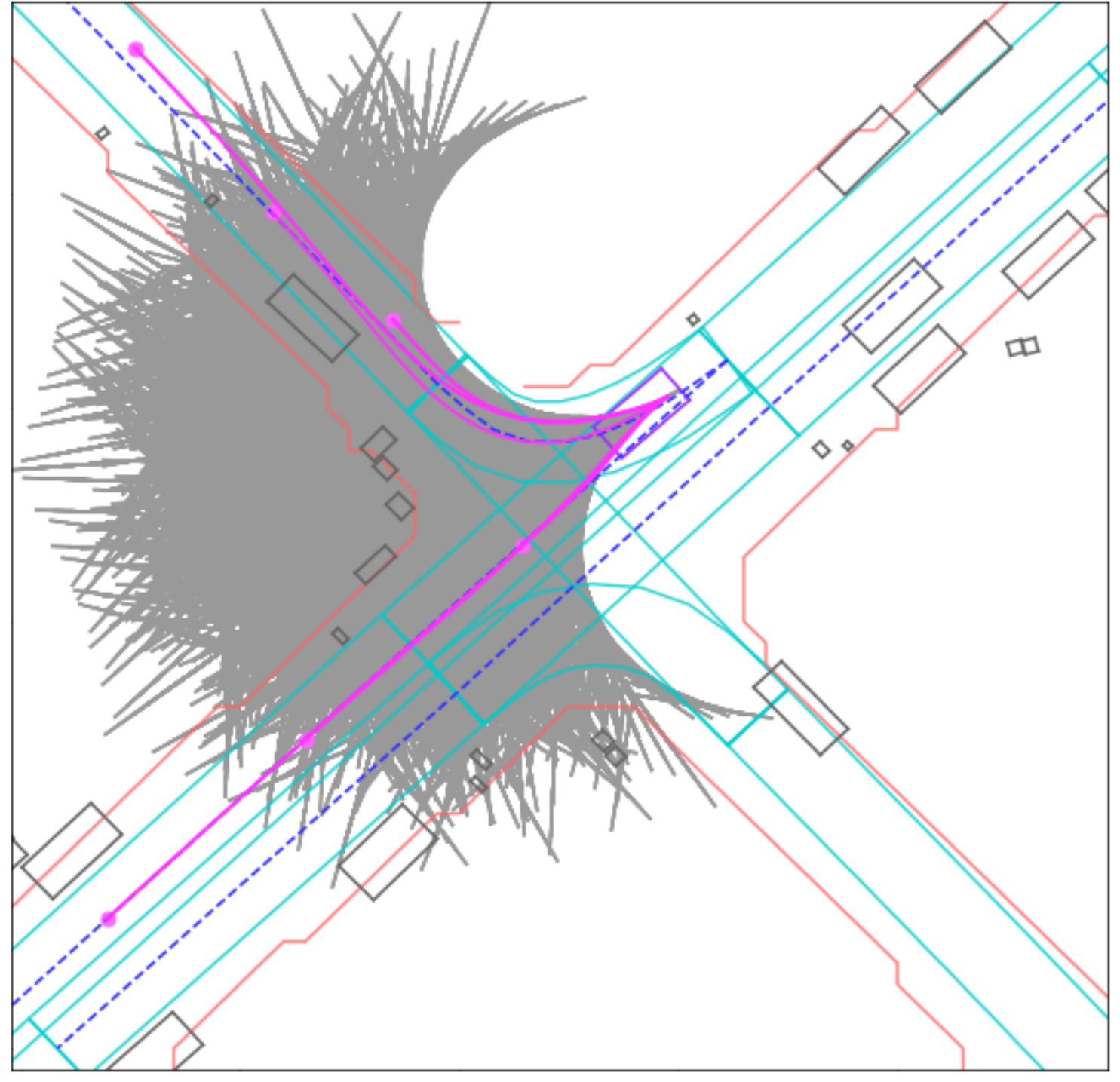}
	\caption{Illustration of using semantic-level action to guide the exploration of the action space. Even with coarsely discretized control actions (e.g., three discretized longitudinal velocities and three lateral accelerations), the action space to be explored (\textit{gray}) of a POMDP-based planner is huge during a multi-step look-ahead search, while most of the space explored is of low-likelihood. The situation is much worse when considering the multi-agent setting. After incorporating semantic-level action (e.g., pursuing center-lines with different aggressiveness) using predefined controllers, the exploration (\textit{purple}) can be guided using domain knowledge. Although much of the exploration is pruned, the resulting exploration still faithfully captures the potential high-level decisions.~\label{fig:semantic_action}}
\end{figure}

It is notable that human drivers do not bear a fine-grained lattice in their mind. Instead, they tend to reason about \textit{only a few} long-term \textit{semantic-level actions} given common driving knowledge (e.g., lane keeping, yielding, overtaking, etc.) which makes the decision-making process extremely efficient. By utilizing semantic-level actions, we can directly sample long-term high-likelihood trajectories for action branching, as shown in Fig.~\ref{fig:semantic_action}. Another key issue of POMDPs is that the intention of other agents is not directly observable. The POMDP requires sampling over the intentions of other agents, which we refer to branching in the observation space. The number of required samples scales exponentially with respect to the number of agents, which is another source of the efficiency problem.

The motivation of guided branching is to mimic the decision-making process of human drivers and utilize domain knowledge for efficient exploration. In this paper, we borrow the idea from MPDM~\cite{cunningham2015mpdm,galceran2017mpdmar} and push our behavior planner one step forward in the following three aspects:
\begin{enumerate}
	\item \textbf{Flexible driving policy}: We reduce the decision space of the original POMDP by guided branching in the action space. Compared to MPDM, the proposed method considers \textit{multiple future decision points} rather than using a single semantic action in the whole planning horizon, thereby obtaining more flexible motions.
	\item \textbf{Efficient policy evaluation}: Instead of directly sampling all possible combinations of agent intentions, which leads to exponential complexity, we propose a mechanism to pick out the potentially risky scenarios conditioned on the ego policy to reduce computational expense during the policy evaluation.
	\item \textbf{Enhanced risk handling in real traffic}: The proposed planner provides an implementation of a new forward simulation model with a safety mechanism, which can reduce the risks caused by the inconsistency between real traffic participants and assumed models. 
\end{enumerate}

\subsection{Guided Action Branching}\label{sec:bp_guided_action}
To guide the exploration in the large space, we introduce semantic-level actions. An exemplary demonstration is provided in Fig.~\ref{fig:semantic_action}. Essentially, the semantic action is a high-level action which can be directly interpreted using human senses, such as \textit{lane-change} and \textit{deceleration}. The semantic action contains multiple small steps and can be executed in a closed-loop manner, it generates a primitive action (i.e., steering angle and longitudinal acceleration) at each step according to a predefined controller. Compared to the primitive action, employing semantic actions constrains the exploration within a higher likelihood region and can generate human-like motion naturally.
Moreover, the duration of semantic actions can be much longer (up to several seconds), which can effectively reduce the height of the belief tree to a relatively small number while obtaining a large enough planning horizon.

Mathematically, the assumption above introduces an additional variable $\phi^i_t\in\Phi^i$ into the driver model, which represents the semantic action for agent $i$ at time $t$, and $\Phi^i$ denotes a set of discretized predefined semantic-level actions for agent $i$. Note that the semantic action set can be different depending on the agent type. For example, the ego semantic set $\Phi^0$ can be larger since more actions lead to more diverse and flexible driving behavior, while the size of $\Phi^{i\neq0}$ for the other agents is designed to be relatively small due to the complexity concern. Moreover, even $\Phi^{i\neq0}$ for different types of road-user (e.g., pedestrians, cyclists, and vehicles) can be different as well. At each step, a semantic action $\phi^i_t$ produces a primitive action $a^i_t$ based on the previous \textit{joint} state $x_{t-1}$ according to a predefined controller $p(a^i_t|x_{t-1}, \phi^i_t)$. This reflects the fact that the mapping from semantic actions to primitive actions takes the surrounding driving context into account, which preserves the closed-loop property of the forward simulation.

To incorporate the semantic action, the joint belief update (Eq. \ref{eq:belief_update}) can be represented as $b_t(x_{t})=p(x_{t}|z_t, \phi_t, b_{t-1})$ by substituting the primitive action with the semantic action, and the joint state transition becomes $T(x_{t-1},\phi_{t},x_t)$. Here, we point out again that the only controllable element is the ego semantic action ($\phi_t=\phi^0_t$), then, we can expand it similar to Eq. \ref{eq:transition_decomp} and obtain the new joint state transition:
\begin{align}
    p(x_{t}|x_{t-1},\phi_t)\approx & 
    \int\limits_{\mathcal{A}^0}\underbrace{p(x^0_t|x^0_{t-1}, a^0_t)}_{\text{ego state transition}}\underbrace{p(a^0_t|x_{t-1}, \phi^0_t)}_{\text{ego controller}}da^0_t \nonumber \\
    & \prod^{N}_{i=1}\int\limits_{\mathcal{A}^i}
    \underbrace{p(x^i_{t}|x^i_{t-1},a^i_{t})}_{\text{$i$-th agent's transition}}
    \underbrace{p(a^i_{t}|x_{t-1})}_{\text{driver model}}da^i_t.
    \label{eq:new_transition}
\end{align}

Note that during the planning process, the ego semantic action $\phi^0_t$ is specified by the policy that the planner is designed to explore. Similarly, we introduce semantic actions for other agents $\phi^i_t\in\Phi^i$ and re-formulate the driver model in Eq.~\ref{eq:new_transition} as
\begin{equation}
    p(a^i_t|x_{t-1}) = \sum_{\phi^i_t\in\Phi^i}\underbrace{p(a^i_t|x_{t-1},\phi^i_t)}_{\text{agent controller}}
    \underbrace{p(\phi^i_t|x_{t-1})}_{\text{new driver model}},
    \label{eq:agent_driver_model}
\end{equation}
wherein $p(\phi^i_t|x_{t-1})$ reflects the decision process of the the $i$-th agent, which maps the driving context to its high-level action.

\begin{figure}[t]
	\centering
	\includegraphics[width=0.4\textwidth]{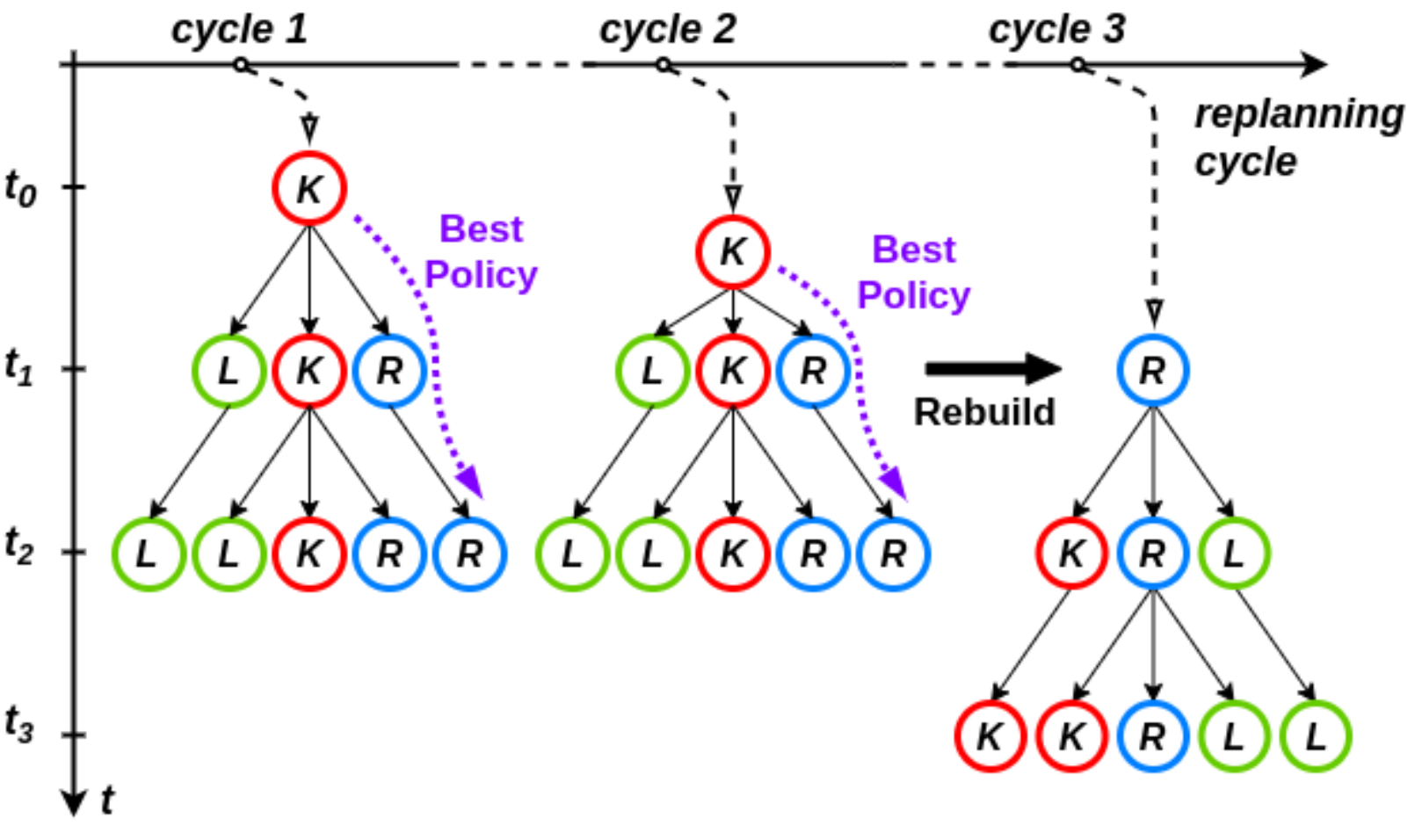}
	\caption{Illustration of the proposed DCP-Tree and the rebuilding process after its \textit{ongoing} action changes. Suppose there are three discrete semantic-level actions $\{K, L, R\}$, denoting ``keep lane", ``left lane change" and ``right lane change", respectively. The \textit{ongoing} action for the left and middle tree is $K$, and the best policy is the dashed \textit{purple} path. After executing $K$, the \textit{ongoing} action switches to $R$, and the DCP-Tree is updated to the right one. Each path only contains one change of action.}\label{fig:dcp_tree_mech}
	\vspace{-0.6cm}
\end{figure}

Similar to MPDM, we prune the belief tree in the action space by directly predefining the decision space within a limited number of policies. One step further, we extend the decision point in the time domain by introducing \textit{domain-specific closed-loop policy tree} (DCP-Tree). The nodes of the DCP-Tree are predefined semantic actions associated with a certain time duration and the directed edges of the tree represent the execution order in time. The DCP-Tree starts from an \textit{ongoing} action $\hat{\phi}$, which is the executing action from the best policy in the last planning cycle. Every time we enter a new planning episode, the DCP-Tree is rebuilt by setting $\hat{\phi}$ to the root node. Note that the number of possible policy sequences scales exponentially w.r.t. the depth of the tree. To overcome this, the DCP-Tree is expanded according to a predefined strategy, which comes from the observation that, as human drivers, we typically do not change the driving policy back and forth in a single decision cycle. Therefore, from the~\textit{ongoing} action, each policy sequence will contain at most one change of action in one planning cycle, as shown in Fig.~\ref{fig:dcp_tree_mech}, while the back-and-forth behavior is achieved by re-planning. Note that the size of the leaf nodes in the DCP-Tree is $O[(|A|-1)(h-1) + 1]$, $\forall h > 1$, which grows linearly with respect to the tree height $h$. For more details about the DCP-Tree, we refer readers to~\cite{ding2020eudm}.

Without loss of generality, we define the policy $\pi\in\hat{\Pi}$ of our behavior planner as a sequence of semantic-level actions which are executed one by one. Once the DCP-Tree is built, the whole decision space $\hat{\Pi}$ of the ego agent is determined. Candidate policies can be obtained by traversing all paths on the DCP-Tree from the root node to all leaf nodes. Therefore, the behavior planning is simplified as selecting the best policy with maximum utility from the candidate policy set. Note that since the evaluations of each policy are independent of each other, we can implement the planner in a parallel fashion without additional effort.

\subsection{Guided Observation Branching}\label{sec:bp_guided_observation}
To evaluate the utility of the policy, a common practice is to use the Monte Carlo-based method. The basic idea is to sample the starting state in the initial belief and use a black-box simulator which mimics the transitions and observations to generate the following states. Then, the belief of each node can be approximated using a particle filter as long as we have enough samples~\cite{silver2010pomcp}. However, sampling over a high-dimensional space lacks efficiency, and modeling randomness in the simulation model makes it even worse. In this work, aiming at real-time applications, we follow the paradigm of black-box simulation and do further simplification for the belief update process. We assume the ego state $x^0_t$ is fully observable and the observations $z_t$ (i.e. tracklets) are noise-free. This is a fair assumption since the ego motion and moving objects information can be estimated with high precision using state-of-the-art methods~\cite{qin2017vins, li2019stereorcnn}. Then, we let the agent policy $p(\phi^i_t|x_{t-1})$ in Eq.~\ref{eq:agent_driver_model}, i.e., the mapping from joint state to agent's semantic action, be deterministic by assuming the semantic action is fully determined by the hidden states, such as intention, and thereby the randomness of the agents' actions are converted into the belief state. As for the the transition model $p(x^i_t|x^i_{t-1},a^i_t)$, we also use a deterministic kinematic model while the control noises can be injected to reflect the inaccuracy of the assumed model if the computational budget is sufficient. After the simplifications above, the only source of uncertainty is the hidden states of the other agent and their update processes after receiving observations.

Based on the understanding of how action branching is guided and implemented, the remaining problem is to estimate the semantic actions $\phi_t^{i\neq0}$, or going one step further, the hidden states of other agents, which are partially modeled in the observation model $p(z^i_t|x^i_{t})$. Different from our previous work~\cite{ding2020eudm}, for other traffic participants, we adopt the assumption that the hidden states of other agents are not updated in the forward simulation, which means that the semantic actions of other agents are fixed. The reason is that once we apply the belief update in the simulation, the policy evaluation will suffer from heavy computation with exponential complexity ($O(|\mathcal{Z}|^d)$), which is unfavorable in real-time applications. 
And, in practice, we find the assumption for other traffic participants leads to satisfactory future anticipation, as shown in Section~\ref{sec:experimental_results}. 
Finally, we estimate the belief of hidden states of other agents in the beginning, and the prediction result is used as the initial belief. We obtain a scenario by sampling over the initial belief and realize it by conducting the deterministic forward simulation:
\begin{align*}
    b_{t_H}(x_{t_H})=b_{t_0}(x_{t_0})
    \prod^{t_H}_{t=t_0}&\Big\{
    \underbrace{p(x^0_t|x^0_{t-1},a^0_t)p(a^0_t|x_{t-1},\phi^0_t)}_{\text{ego state propagation}}\\
    & \prod^{N}_{i=1}
    \underbrace{p(x^i_t|x^i_{t-1},a^i_t)p(a^i_t|x_{t-1},\phi^i_0)}_{\text{other agents' state propagation}}\Big\},
\end{align*}
where the ego semantic action $\phi^0_t$ is provided by the current policy. $\phi^{i\neq0}_t$ only depends on the initial belief $b_{t_0}$ and remains fixed during the forward simulation. Note that although we employ multiple simplifications, the forward simulation is still a closed-loop process since actions of all agents depends on the simulated state of the previous step $x_{t-1}$, retaining the interaction-awareness. As a result, the total utility of the policy is the weighted sum of the sampled scenarios, while the weights are exactly the value of the initial belief.

In this work, the semantic action for the other agent is represented by the action of pursuing a particular center-line. Therefore, estimating $\phi_t^{i\neq0}$ is equivalent to estimating the probability distribution over the candidate center-lines that a certain agent is pursuing. In highway environments, this task is straightforward due to the simple lane structure, and we adopt a rule-based predictor in our previous work~\cite{ding2020eudm}. In urban environments, this task is more challenging due to more complex road structures. Essentially, in urban environments we are facing a varying number of candidate center-lines depending on the scene. To deal with this, we propose a lightweight classification network for estimating the probabilities while supporting a varying number of classes. The details of the proposed learning-based intention estimator are elaborated in Section~\ref{sec:impl_intention}.

\begin{figure}[!bt]
	\centering
	\includegraphics[width =0.4\textwidth]{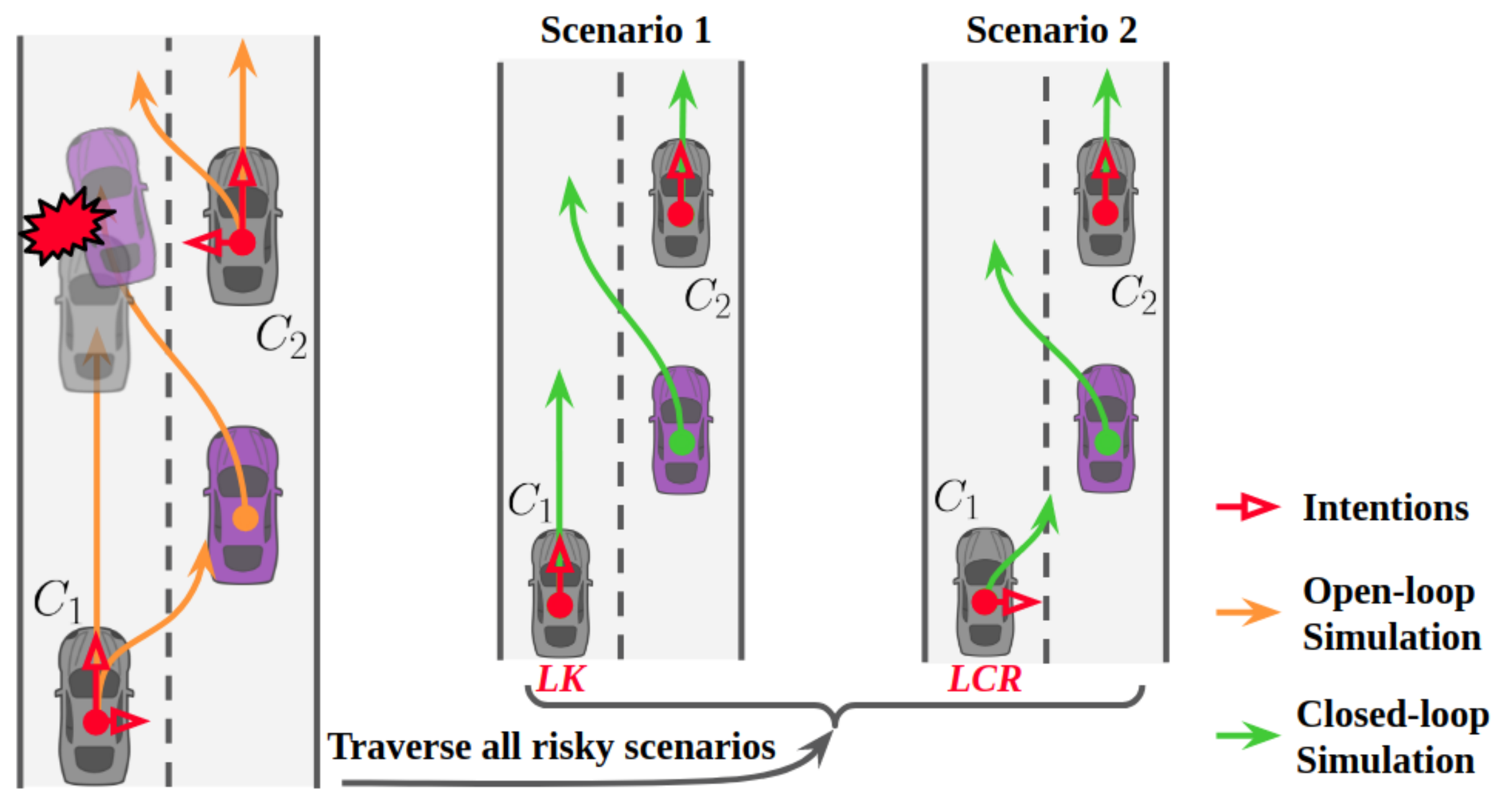}
	\caption{Illustration of the CFB mechanism. Considering a case that the ego vehicle (purple) is considering a left lane-change. We use the open-loop simulation to get all possible future trajectories for all agents. Then, for each surrounding agent, we check whether there would be potential risks according to the simulated trajectories. For the critical agent ($C_1$), we unfold all combinations of the possible intentions, and for the vehicle with low risk ($C_2$), we directly employ the intention with maximum probability. As a result, the policy evaluation turns into an evaluation of the two scenarios. Interested readers may refer to~\cite{ding2020eudm} for more details.}\label{fig:cfb_toy_case}
	\vspace{-0.6cm}
\end{figure}

After obtaining the initial belief, the remaining problem is to determine $\phi_t^{i\neq0}$ based on the estimated probabilities. Since each traffic participant potentially has multiple choices (e.g., multiple candidate center-lines), we define one combination of intentions for all the traffic participants as one \textit{scenario}. Intuitively, there is an exponential number of scenarios and the probability for each scenario can be derived from the joint distribution of the estimated probabilities for each participants.\footnote{We assume each traffic participant makes decisions independently at each time instance.} It is too expensive to examine all the possible scenarios when the agent number is large. In~\cite{ding2020eudm}, we propose using conditional focused branching (CFB) for this task. Essentially, the goal of CFB is to find the intentions of nearby vehicles that may lead to risky outcomes with as few branches as possible. The term ``conditional'' means conditioning on the ego policies. The motivation comes from the observation that the attention of the human driver for nearby vehicles is biased differently when intending to conduct different maneuvers. For example, a driver will pay more attention to the situation in the left-hand lane rather than the right-hand one when he intends to make a left lane-change. As a consequence, by conditioning on the ego policy sequence, we can pick out a set of relevant vehicles w.r.t. the ego policy. A toy example is provided in Fig.~\ref{fig:cfb_toy_case}. The selection process is currently based on the rule-based safety assessment, which is identical to that in~\cite{ding2020eudm}. We point out that the critical agent selection can be implemented using learning-based methods based on real-world data. However, in practice, we find the proposed method can identify many risky cases despite its simple design.

\begin{figure}[t]
	\centering
	\includegraphics[width =0.42\textwidth]{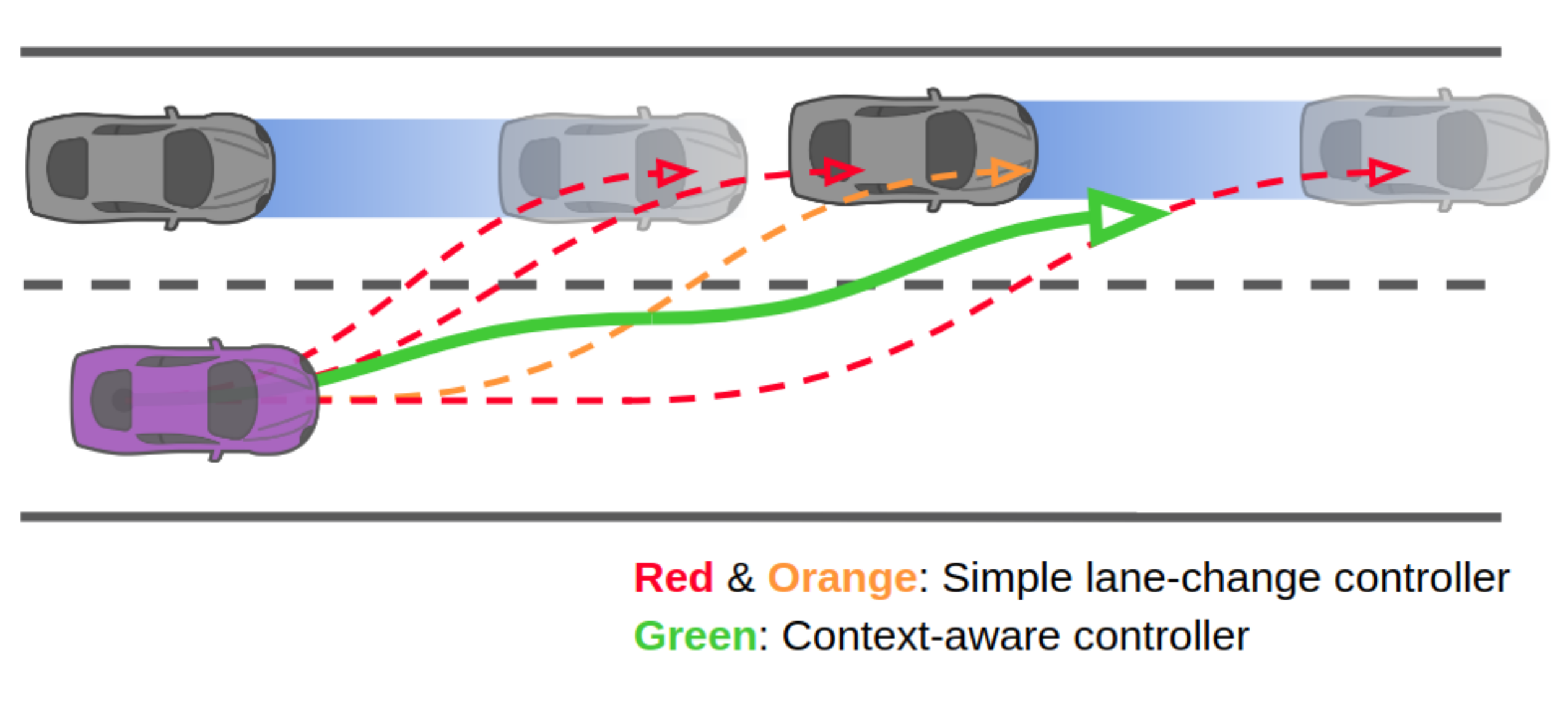}
	\caption{Illustration of the difference between the simple controller and \textit{context-aware}  controller in the action branching. The lane-changing manoeuvres launched at different timestamps are realized by a simple pure pursuit controller (dashed curves). However, the simple controller does not \textit{actively} pursue lane-changing, resulting in failure cases (red) or risky cases (orange), which is very inefficient. We find that it is beneficial to utilize the \textit{context-aware} controller to achieve an advanced driving style, e.g., gap finding and velocity adaption. The resulting motion automatically adapts to different traffic configurations (green) with a higher success rate.}\label{fig:advaned_policy}
	\vspace{-0.4cm}
\end{figure}

\subsection{Multi-agent Forward Simulation}\label{sec:bp_forward_integration}

As mentioned above, a \textit{scenario} is realized by the forward simulation, which propagates all agents in the environment simultaneously in an interactive manner. During the multi-agent forward propagation, all trajectories of the ego vehicle and other agents are generated. It is worth noting that in many existing works (e.g.,~\cite{montemerlo2008junior,urmson2008autonomous,ziegler2014making,ajanovic2018search,hubmann2016generic,wei2014behavioral,zhan2017spatially,fan2018baidu}), this type of task is conducted in an independent trajectory prediction module. However, after decoupling prediction and planning, it is problematic to account for the impact of the ego vehicle's future motion on the other agents. By contrast, our method couples the prediction inside the behavior planning where each potential scenario considered by the planner has its own corresponding future anticipation. Essentially, at each time step in the planning horizon, each agent observes the latest states of other agents and obtains model-based control using a predefined driver model $p(a^i_t|x_{t-1})$. Based on the control, the agent's state is pushed one step forward following the state transition $p(x^i_{t}|x^i_{t-1},a^i_{t})$ of the agent.

In ~\cite{ding2020eudm} we designed a \textit{simple model-based} driver model by using the intelligent driver model (IDM)~\cite{treiber2013traffic} for velocity control and pure pursuit~\cite{coulter1992implementation} for steering control.
Different from our previous work, we differentiate the controlled vehicle from other traffic participants. For the controlled vehicle, we design a \textit{context-aware} controller for flexible and safe ego behaviors, while for other traffic participants we stick to a \textit{simple model-based} controller. The proposed \textit{context-aware} controller also contains a velocity controller and a steering controller, however, compared to the previous \textit{simple model-based} approach, the new controller can generate more realistic maneuvers, especially in the lane change process.
The reason that we utilize the \textit{context-aware} controller is that there is considerable potential in the policy design for realizing the semantic-level actions, which can facilitate enforcing advanced driving styles during action branching. An intuitive example is provided in Fig.~\ref{fig:advaned_policy}, in which we show how lane changing is conducted. By introducing the new controller, the solution space is enlarged and the resulting decision can adapt to the traffic configuration in a more flexible way. Benefiting from the \textit{context-aware} controller, the number of action sequences needed is reduced significantly, and the success rate of the forward simulation is greatly improved, which further enhances the performance and efficiency of branching. Actually, the implementation of such controllers has been well studied in the traffic simulation literature and we detail our implementation in Section~\ref{sec:bp_forward_integration}.

Although the forward simulation model seems quite straightforward, we find it is actually accurate enough for anticipating the future. To our surprise, our lightweight design outperforms the two state-of-the-art trajectory prediction methods listed in Section~\ref{sec:experimental_results} in terms of multi-modal prediction. Moreover, our design provides the possibility of coupling prediction and planning, which can not be achieved in traditional trajectory prediction methods. From the perspective of prediction, our multi-agent forward simulation is like a ``conditional'' prediction, which conditions the prediction process on the decision of the controlled vehicle. \cite{rhinehart2019precog, song20pip} show that the predictor can achieve much better performance by considering the potential future plan of the agent.

\subsection{Safety Mechanism}\label{sec:safety_mechnism}
The foregoing discussion does not answer one important question: what if the real-world driver deviates from the predefined model?
In practice, we observe that this issue often leads to underestimation of risk in decision-making, since real-world drivers are often stochastic and noisily rational. In traditional methods, this issue is dealt with by leaving sufficiently large safety margins~\cite{werling2012mp,naumann2020irl}, which restricts the flexibility in highly interactive environments. To deal with this issue, we propose a safety mechanism with two levels.

Firstly, we enhance the safety of the forward simulation model by embedding a safety module in the \textit{context-aware controller} which can automatically ensure the control output is safe or \textit{no-fault} to some extent. In terms of responsibility and safety, Shai~\textit{et al.}~\cite{shalev2017formal} defines the responsibility-sensitive safety (RSS), which provides a mathematical model for safety lateral and longitudinal distances in various driving situations and the \textit{proper response} in dangerous situations (i.e., when safety distances are not satisfied). For each simulation step, we check whether the simulated state for the controlled vehicle is RSS-safe. If not, we check whether the control signal provided by the \textit{context-aware} controller follows the criterion of \textit{proper response} defined by the RSS model. If still not, we generate a safe control signal obeying the proper response and override the control output. We refer interested readers to~\cite{shalev2017formal} for a detailed description of RSS. 

Secondly, we strengthen the robustness of the decision layer by setting the safety criterion in the policy selection. We are motivated by the reason why human drivers may perform an aggressive cut-in maneuver: for one thing, human drivers are aware of the possibility that other drivers may react to their insertion, for another, experienced human drivers have a safety evaluation process in their mind, namely, even in the case that other drivers are not cooperative, they still have the maneuverability to cancel their lane-change if they cannot complete it. Benefiting from our policy design, this feature can be achieved via the so called \textit{backup plan}, which is naturally possessed by our behavior planner. Since our behavior planner evaluates multiple policies in one planning cycle, we can take any other policy that successfully fulfils the forward simulation and the safety check as a backup plan. In order to be more targeted, we handcraft a priority for the backup policy selection. For example, the backup plan for the ``lane-change'' policy should be ``lane change cancelled''. We can find the corresponding backup policies and check whether there is at least one feasible plan. If so, the planned policy can be executed. If there are no safe policies at all, meaning that the behavior planning has failed, a warning signal will be displayed, and the low-level active protection, such as autonomous emergency braking (AEB), will be triggered.

\subsection{Policy Selection}
After the guided branching in both the action space and observation space, as well as the approximation of the transitions and observations, the behaviour planner can be simplified into a limited number of policy evaluation problems. For each policy, we calculate the weighted sum of the reward of each scenario by evaluating the planned behavior and the simulated trajectory. The reward function consists of three parts, namely, efficiency, safety and navigation, which are elaborated in Section~\ref{sec:policy_eval}. Finally, the flow of our behavior planner is described in Algo.~\ref{algo:policy_selection}. Evaluation for each policy sequence can be carried out in parallel (Line 5 to 11). Each critical scenario selected by CFB is examined by closed-loop forward simulation (Line 8 to 10) in parallel, and each policy is evaluated (Line 11) using the reward function, with the best policy selected (Line 13).

\begin{figure}[t]
	\removelatexerror
	\begin{minipage}{.48\textwidth}
		\begin{algorithm}[H]\label{algo:policy_selection}
			\caption{Process of behavior planning layer}
			Inputs:
				Current states of ego and other vehicles $x$;
				\textit{Ongoing action} $\hat{\phi}$;
				Pre-defined semantic action set $\Phi$;
				Planning horizon $H$\;
			$\mathfrak{R}\leftarrow\emptyset$; $\slash\slash$ set of rewards for each policy\;
			$\Psi \leftarrow  \func{UpdateDCPTree(\Phi,\hat{\phi})}$; $\slash\slash$ DCP-Tree $\Psi$\;
			$\hat{\Pi}\leftarrow\func{ExtractPolicySequences}(\Psi)$\;
			\ForEach{${\pi}\in\hat{\Pi}$}
			{
				$\Gamma^{{\pi}}\leftarrow\emptyset$; $\slash\slash$ set of simulated trajectories\;
				$\Omega \leftarrow\func{CFB}(x,{\pi})$; $\slash\slash$ set of critical scenarios\;
				\ForEach{$\omega \in \Omega$}{
					$\Gamma^{{\pi}}\leftarrow\Gamma^{{\pi}}\cup\func{SimulateForward}(\omega,{\pi},H)$\;
				}
				$\mathfrak{R}\leftarrow\mathfrak{R}\cup\func{EvaluatePolicy}({\pi},\Gamma^{{\pi}})$\;
			}
			${\pi}^*,\hat{\phi}\leftarrow\func{SelectPolicy}(\mathfrak{R})$\;
		\end{algorithm}
	\end{minipage}
	\vspace{-0.6cm}
\end{figure}

\section{Trajectory Generation with Spatio-temporal Semantic Corridor}\label{sec:ssc_traj}

As stated in Section~\ref{sec:bp}, the generated final decision $\mathcal{D}_t = [x_{t+1}, x_{t+2},\ldots, x_{t+H}]$ from the behavior planning layer contains a sequence of states for both the ego vehicle and the other agents. The remaining problem is to generate a smooth, safe and dynamically feasible trajectory which faithfully follows the decision $\mathcal{D}_t$. In this paper, we follow the formulation of our previous work~\cite{ding2019safe} that characterize the local solution space around $\mathcal{D}_t$ using the spatio-temporal semantic corridor (SSC), and then use an optimization-based formulation for trajectory generation within the SSC. For simplicity, we omit the detailed formulations here, and we refer interested readers to~\cite{ding2019safe}.

Due to the hierarchical structure of our planning system, the formulations and objectives of the behavior and motion layer are different, resulting in the potential discrepancy between the two layers and leading to inconsistent plan. Different from \cite{ding2019safe}, in addition to minimizing the jerk along the trajectory, we incorporate the mean squared error of the trajectory to the reference state at the corresponding time, which represents the \textit{similarity} between the trajectory and the ego decision in $\mathcal{D}_t$. Specifically, the cost $J^{\sigma}_j$ of the $j$-th segment on dimension $\sigma$ can be written as
\begin{align}
	J^{\sigma}_j = w^{\sigma}_s\cdot\int_{t_{j-1}}^{t_j} &\left(\frac{d^3 f^{\sigma}_j(t)}{dt^3} \right)^2 dt\nonumber\\
	& + w^{\sigma}_{f}\cdot\frac{1}{n_j}\sum^{n_j}_{k=0}\left(f^{\sigma}_j(t_k)-r^{\sigma}_{jk}\right)^2,
\end{align}
where $w^{\sigma}_s$ and $w^{\sigma}_f$ denote the weight for the \textit{smoothness} cost and the \textit{similarity} cost, respectively. $r^{\sigma}_{jk}$ denotes the $k$-th reference state on dimension $\sigma$ generated by the ego-simulated states corresponding to the $j$-th segment, and $t_k$ is the timestamp of $r^{\sigma}_{jk}$. The number of reference states $n_j$ in the $j$-th segment is determined by the result of cube inflation. Note that we can further simplify the cost function and rewrite it in a quadratic formulation (ignoring constant terms):
\begin{align*}
    J^{\sigma}_j &= \mathbf{p}_j^{\text{T}}(w^{\sigma}_s\mathbf{Q}_s + w^{\sigma}_f\mathbf{Q}_f)\mathbf{p}_j + w^{\sigma}_f\mathbf{c}^{\text{T}}\mathbf{p}_j\\
    &=\frac{1}{2}\mathbf{p}_j^{\text{T}}\hat{\mathbf{Q}}\mathbf{p}_j+ \hat{\mathbf{c}}^{\text{T}}\mathbf{p}_j,
\end{align*}
where $\mathbf{Q}_s$ and $\mathbf{Q}_f$ are real symmetric matrices related to the smoothness term and similarity term, respectively, and $\mathbf{c}$ is a real vector. The objective function is simple and invariant given different combinations of semantic elements thanks to the SSC, which allows the formulation to easily adapt to different traffic configurations. Finally, the overall formulation can be written as a quadratic programming (QP), which can be solved efficiently using off-the-shelf solvers (such as OOQP~\cite{gertz2003object}). In the case that no feasible solution can be found, the error is fed back to the behavior layer for further reaction.

\section{Implementation Details}\label{sec:implementation}

\subsection{Semantic-level Actions and DCP-Tree}\label{sec:impl_semantic_action}
We consider both lateral and longitudinal actions to obtain a diverse driving policy. For both highway and urban environments, the lateral action set can be defined as the target center-line $c_i$ of the adjacent lanes, since the most common lateral movement for a vehicle is inter-lane-changing among available lanes. Note that the number of lateral actions $N^{lat}_a$ in the set is conditioned on the driving context since the adjacent lanes may be unavailable (due to traffic rules) or just not exist. For longitudinal motion, in order to ensure the continuity and safety, we use a predefined velocity controller rather than directly applying longitudinal acceleration signals. Thus, we define the semantic actions as a set of parameters of the velocity controller obtaining different longitudinal maneuvers. For example, in the intelligent driver model, we can achieve a more aggressive driving style by increasing the desired velocity while decreasing the time headway and minimum spacing. Here, we define the longitudinal semantic action set with $N^{lon}_a=3$ items:~\{\textit{Aggressive, Moderate, Conservative}\}. Finally, we can get $N^{{lon}}_a\times N^{{lat}}_a$ semantic-level actions in total by combining the lateral and longitudinal actions. As for the DCP-Tree, we set the depth of the tree as $5$. We assign a time duration of $1$ s for each semantic-level action, while the closed-loop simulation is carried out with a $0.2$ s resolution to preserve the fidelity. Therefore, the planning horizon of the proposed behavior planner is $5$ seconds.

\subsection{Context-aware Forward Simulation Model}
The goal of the closed-loop simulation is to push the state of the multi-agent system forward while considering the potential interaction. For the proposed \textit{context-aware} driver model, we utilize different lateral and longitudinal controllers for different combinations of semantic actions. Specifically, for lane keeping, the lateral controller follows the pure pursuit controller, while the longitudinal motion is governed by the IDM.
For lane changing maneuvers, the longitudinal and lateral control is coupled since all the vehicles in the current and adjacent lanes need to be considered. We extend the lateral controller as the reactive pure pursuit, and the longitudinal controller as a \textit{gap-informed} velocity controller. 

\begin{figure}[t]
	\centering
	\begin{subfigure}[b]{0.4\textwidth}
		\centering
		\includegraphics[width =0.95\textwidth]{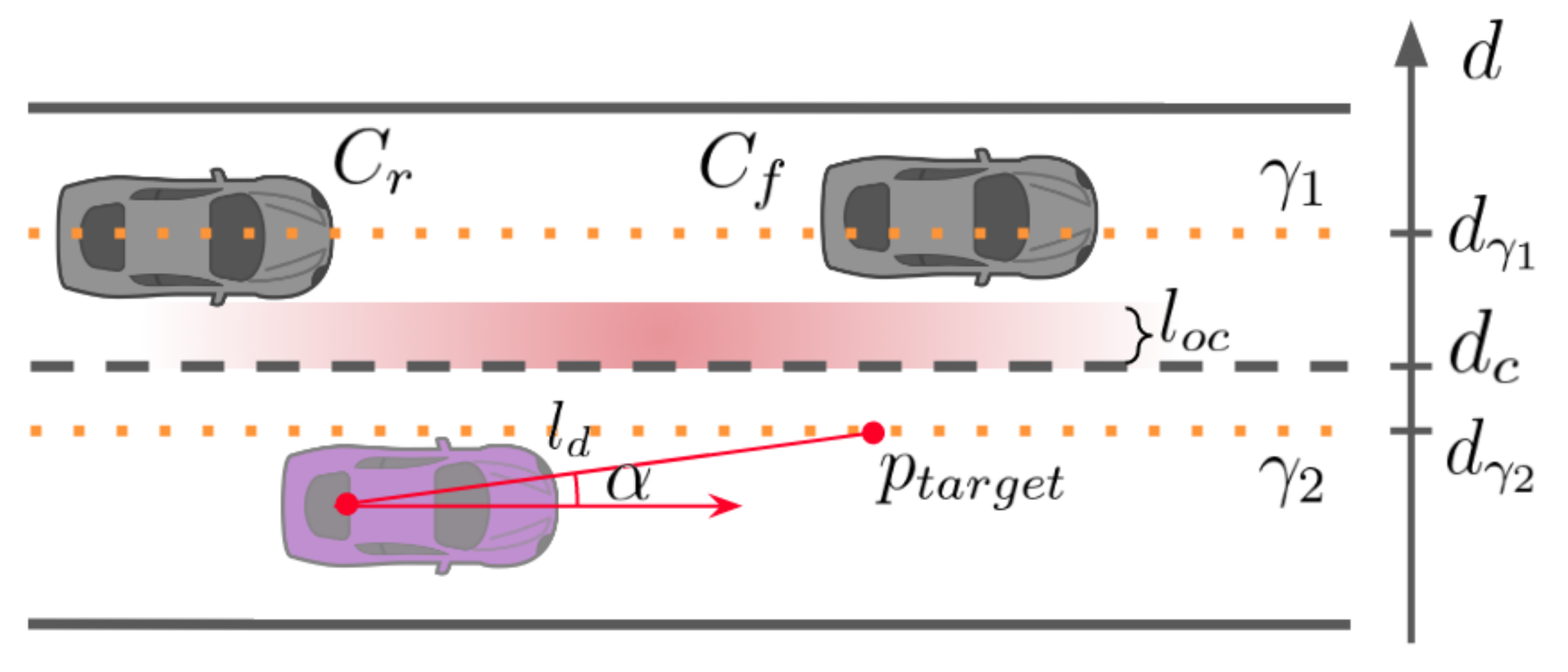}
		\caption{The reactive lateral motion controller for lane-changing with obstacles in the target lane. The origin and alternate reference path are shown in orange, and the minimum clearance between lane markings and obstacles is represented by the red zone. The look-ahead distance $l_d$ and heading error $\alpha$ can be quickly obtained once the target point $p_{target}$ is determined.}\label{fig:sim_pp}
	\end{subfigure}
	\begin{subfigure}[b]{0.4\textwidth}
		\centering
		\includegraphics[width =0.95\textwidth]{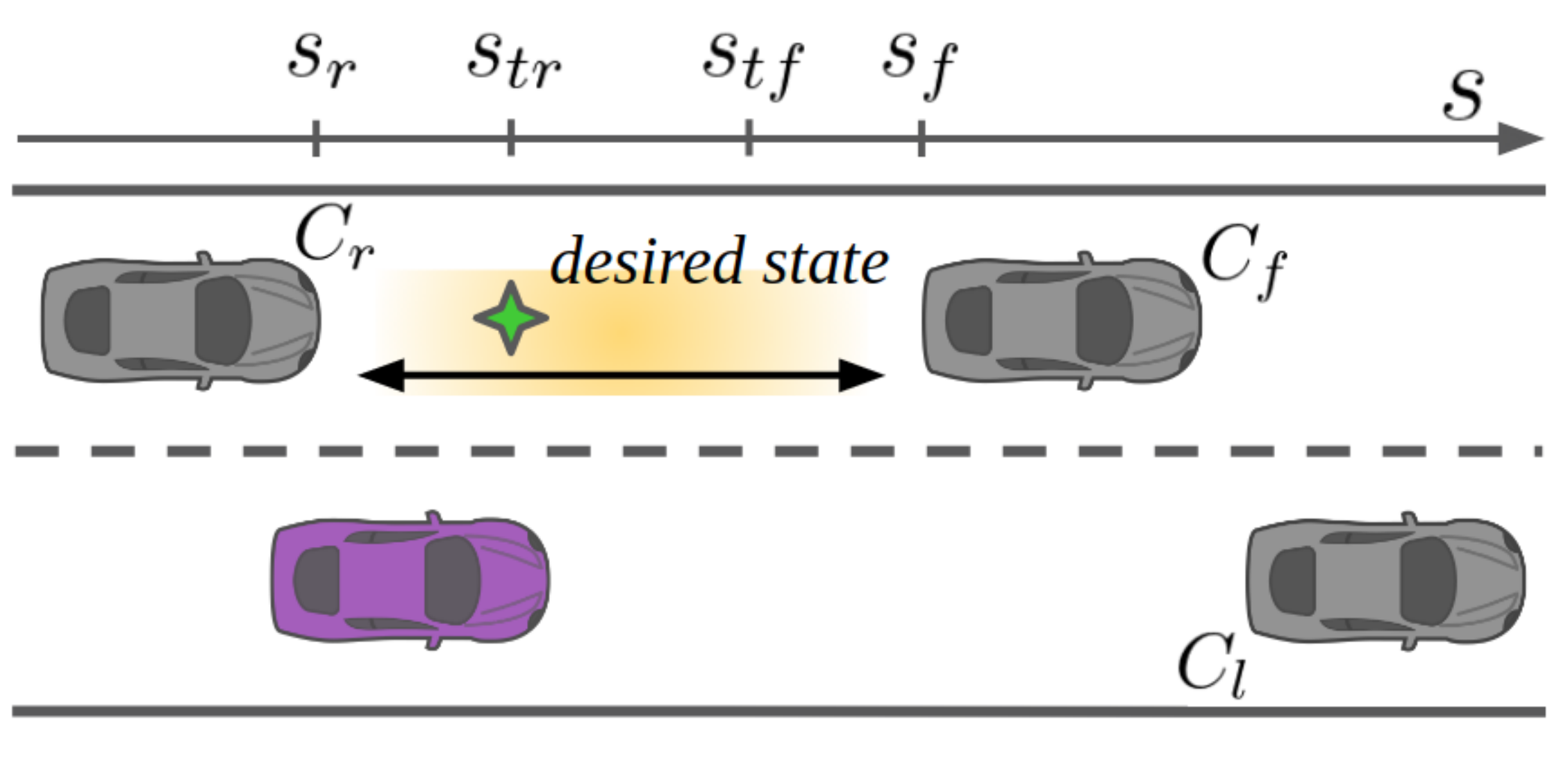}
		\caption{The gap-informed longitudinal motion controller for lane-changing action. We consider three vehicles during lane-changing maneuvers, which are the current leading vehicle $C_l$ and the new leader and follower $C_f$ and $C_r$. The target gap formed between $C_f$ and $C_r$ is shown in yellow, while the desired longitudinal state is marked by the green star.}\label{fig:sim_gap}
	\end{subfigure}
	\caption{Illustration of the proposed forward simulation model. The ego vehicle is shown in purple while the other agents are in gray.}\label{fig:sim_ctrl}
\end{figure}

To better illustrate the proposed forward simulation model, we provide two exemplar scenarios shown in Fig.~\ref{fig:sim_ctrl}. Note that the driving scenarios are presented in the curvilinear coordinate frame, where $s_{(\cdot)}$ and $d_{(\cdot)}$ denote the longitudinal and lateral positions in the path-relative coordinate frame. We brief the related variables in Tab. \ref{tab:fs_variable_desc}, while the important ones are further introduced in the following contents.

For the lateral steering control, we firstly check whether the target lane has enough space for merging. If the adjacent lane is free, the center-line of the target lane $\gamma_1$ is assigned as the reference path of the pure pursuit controller, otherwise (Fig.~\ref{fig:sim_pp}), we use an alternate collision-free path $\gamma_2$ as the reference. The lateral position of $\gamma_2$ can be calculated using
\begin{equation*}
    d_{\gamma_2}=d_c-\frac{W_{ego}}{2}-max(0,l_{safe}-l_{oc}),
\end{equation*}
wherein $l_{oc}$ is the minimum distance between obstacles and the lane marking, and $l_{safe}$ denotes the user-preferred minimum lateral clearance between the ego vehicle and obstacles (see Fig.\ref{fig:sim_pp}). By tracking the alternate reference path $\gamma_2$, the controlled vehicle can slightly move toward the target lane without crossing the lane marking while keeping an appropriate distance from obstacles. The corresponding steering angle can be calculated using the pure pursuit controller $\delta_{ctrl} = \arctan\frac{2L\sin{\alpha}}{l_d}$.

\begin{table}[t]
	\centering
	\caption{Description of the related variables of the proposed forward simulation model.}\label{tab:}
	\begin{tabular}{m{1.5cm} | m{6cm}}
	\toprule
	\textbf{Variables} & \textbf{Description} \\
	\midrule
	$C_l$, $C_r$, $C_f$         & Current leading vehicle, the expected new leader and follower of the ego vehicle after the lane-changing \\ \hline
    $W_{ego}$, $L$              & Width and the wheelbase of the ego vehicle \\ \hline
    $\gamma_1$, $d_{\gamma_1}$  & Target center-line and its lateral position \\ \hline
    $\gamma_2$, $d_{\gamma_2}$  & Alternate reference path and its lateral position \\  \hline
    $d_c$                       & Lateral position of the lane marking between the target and current lanes \\ \hline
    $l_{oc}$                    & Observed minimum distance between the obstacles and the lane marking \\ \hline
    $l_{safe}$                  & User-preferred minimum lateral clearance between the ego vehicle and obstacles \\ \hline
    $p_{target}$                & Look-ahead point for the pure-pursuit controller \\ \hline
    $l_d$, $\alpha$             & Look-ahead distance and the heading error \\ \hline
    $s_{des}$, $\dot{s}_{des}$  & Desired longitudinal position and velocity for the ego vehicle while merging \\ \hline
    $s_{ego}$, $\dot{s}_{ego}$  & Ego longitudinal position and velocity \\ \hline
    $\dot{s}_{pref}$            & Preferred longitudinal velocity assigned by the user and traffic rules \\ \hline
    $s_r$                       & Longitudinal position of the front bumper for the expected rear vehicle $C_r$ \\ \hline
    $s_f$                       & Longitudinal position of the rear bumper for the expected leading vehicle $C_f$ \\ \hline
    $\dot{s}_r$, $\dot{s}_f$    & Observed longitudinal velocities of $C_r$ and $C_f$ \\ \hline
    $s_{tr}$, $s_{tf}$          & Threshold longitudinal positions derived from the safe time headway \\ \hline
	$l_{min}, T_{safe}$         & Minimum spacing and the safe time headway in the longitudinal direction \\
	\bottomrule
	\end{tabular}\label{tab:fs_variable_desc}
\end{table}

For longitudinal control, a gap in the target lane is selected, and a desired longitudinal state is generated according to the states of the new leader $C_f$ and follower $C_r$. The desired longitudinal state $[s_{des}, \dot{s}_{des}]$ can be calculated as
\begin{align*}
    s_{des} &= \min(\max(s_{tr}, s_{ego}), s_{tf})\\
    \dot{s}_{des} &= \min(\max(\dot{s}_r, \dot{s}_{pref}), \dot{s}_f),
\end{align*}
where \ $[\dot{s}_r, \dot{s}_f]$ is the observed longitudinal velocity of $C_r$ and $C_f$, $\dot{s}_{pref}$ is the preferred longitudinal cruise velocity assigned by user and traffic rules, and $s_{tr}$ and $s_{tf}$ are the threshold positions derived from the time headway corresponding to $C_r$ and $C_f$,
\begin{equation*}
    \begin{bmatrix}
        s_{tr} \\
        s_{tf} \\
    \end{bmatrix} = 
    \begin{bmatrix}
        s_r + l_{min} + T_{safe} \dot{s}_r \\
        s_f - l_{min} + T_{safe} \dot{s}_{ego} \\
    \end{bmatrix},
\end{equation*}
where $l_{min}$ and $T_{safe}$ are the minimum spacing and safe time headway. The presented desired state provides us a target for the longitudinal control during the lane-changing maneuver, making the whole process comfortable and safe. Note that even when the bumper-to-bumper distance is small, the desired state still exists and the ego longitudinal state will gradually converge to the target and align with the gap. Furthermore, we point out that the desired state calculation still works when either $C_f$ or $C_r$ is absent by just simply eliminating the related terms in the equations above.

Once the desired longitudinal state is generated, a simple feedback controller is applied to push the ego vehicle toward the desired state. The output of the controller is the longitudinal acceleration $a_{track}$ with the form of
$$a_{track} = K_v(\dot{s}_{des} + K_s(s_{des} - s_{ego}) - \dot{s}_{ego}),$$
where $K_v$ and $K_s$ are the feedback gain of the controller. Note that during the lane-changing process, we also need to consider the current leading vehicle $C_l$ to ensure safety. Here we use the ordinary IDM to get another acceleration $a_{idm}$ for ego-lane distance keeping. Therefore, the output of the velocity controller is defined as $a_{ctrl}\coloneqq\min(a_{track}, a_{idm})$. 

As stated in Section~\ref{sec:bp_forward_integration}, to achieve a higher safety requirement, we double-check the lateral and longitudinal control command using the RSS-safety criterion. Once the ego is RSS-dangerous at the current simulation step, i.e. the situation is not RSS-safe according to the rules introduced in~\cite{shalev2017formal}, feasible longitudinal and lateral limits on acceleration are provided. We compare the control constraints with the output signal ($\delta_{ctrl}$ and $a_{ctrl}$) and check whether the output satisfies the \textit{proper response}. If not, we generate a safe control signal induced by the proper response and override the output. 

For the vehicle model, i.e. the transition model $p(x^i_{t}|x^i_{t-1},a^i_{t})$, all agents are described using the kinematic bicycle model~\cite{kong2015kinematic} for balancing computational efficiency and simulation fidelity.

\subsection{Intention Estimator}\label{sec:impl_intention}

\begin{figure}[t]
	\centering
	\begin{subfigure}[b]{0.45\textwidth}
		\centering
		\includegraphics[width =\textwidth]{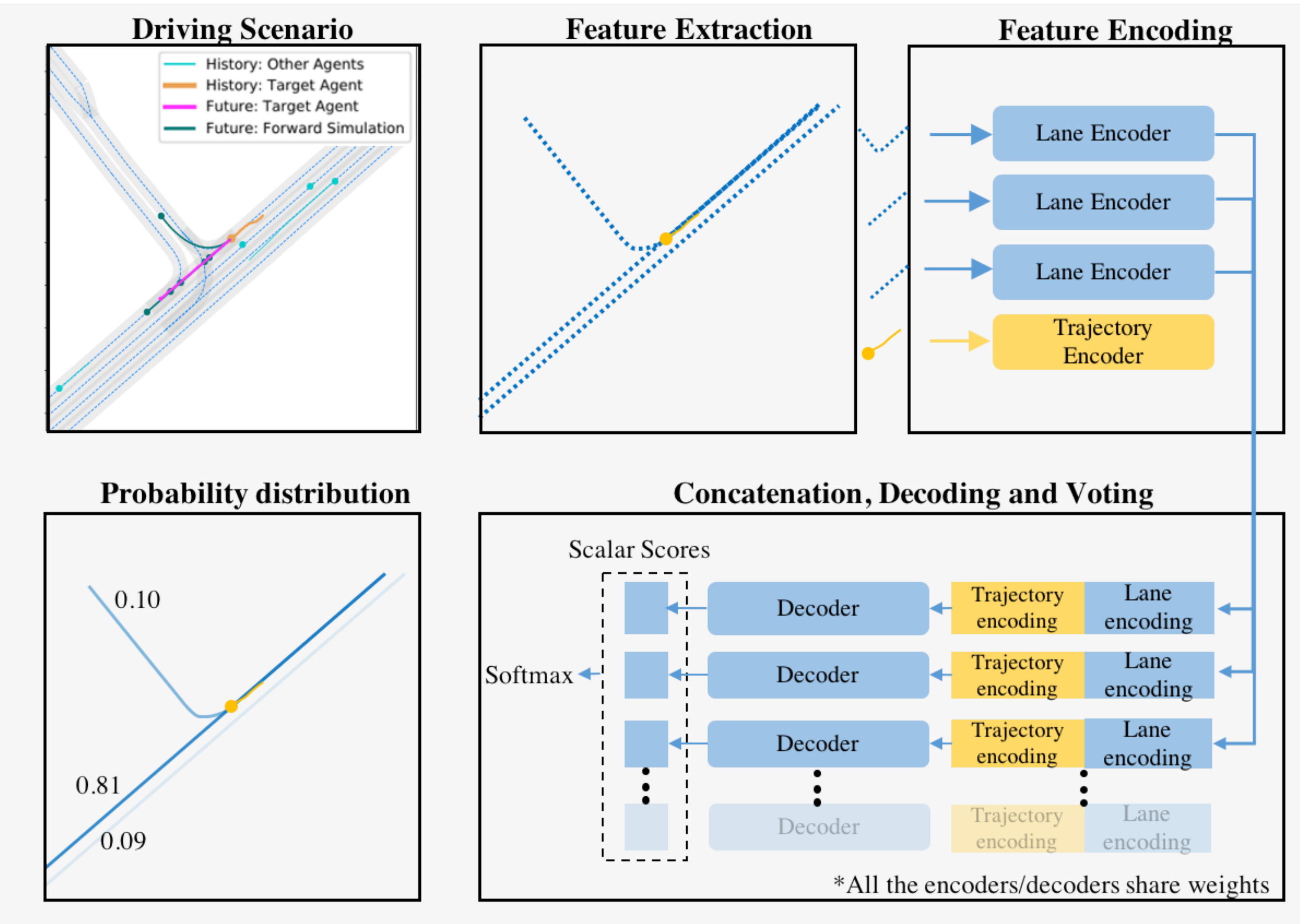}
		\caption{Illustration of our lightweight network for estimating the semantic action (i.e., center-line) of other traffic participants.}\label{fig:network}
	\end{subfigure}
	\begin{subfigure}[b]{0.45\textwidth}
		\centering
		\includegraphics[width =\textwidth]{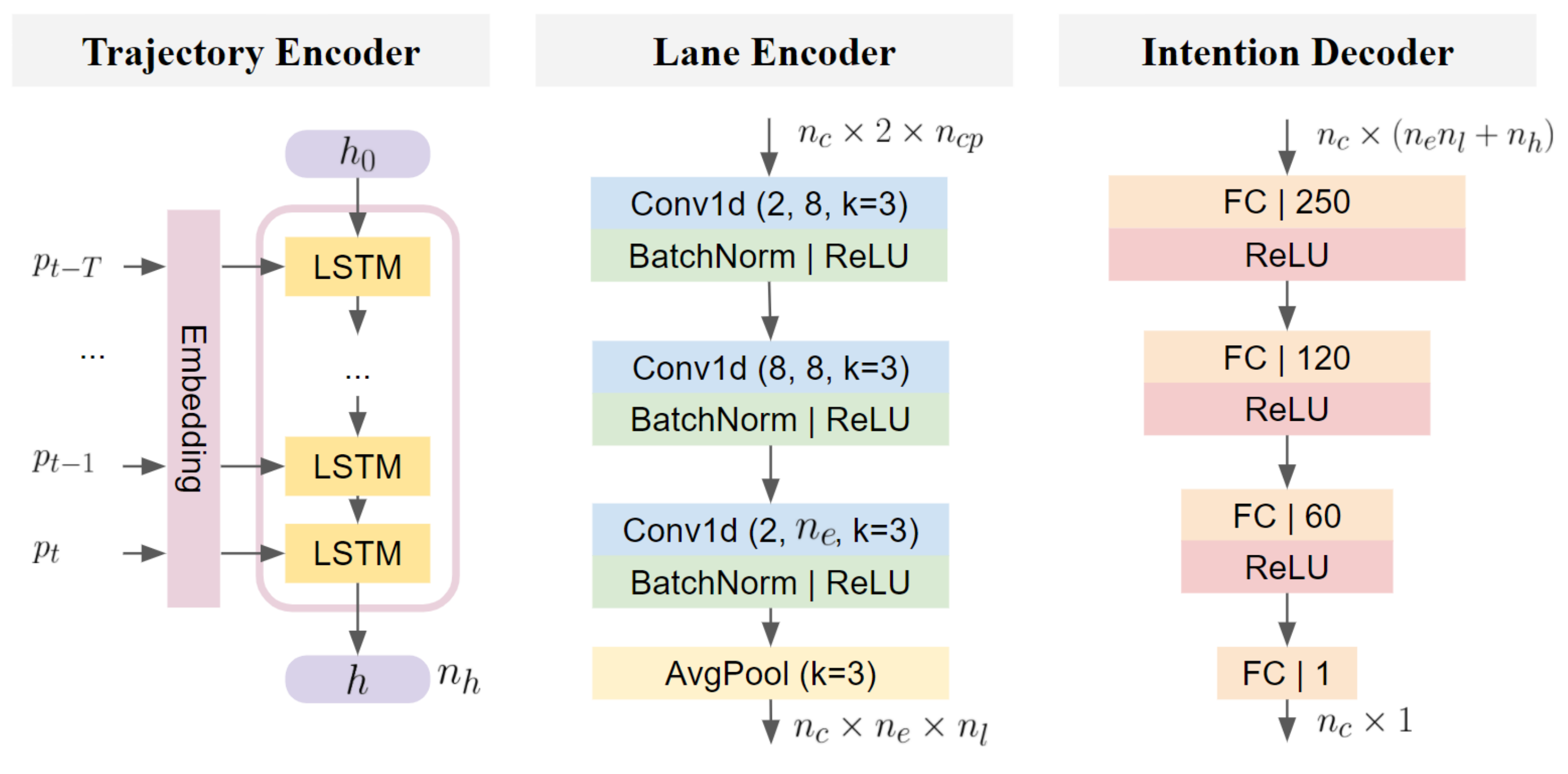}
		\caption{Illustration of the network structure. In the trajectory encoder, $h_0$ is the initial hidden vector, which is a zero vector. In the lane encoder, the size of the input and output channel of the 1-D convolution is presented. $k$ denotes the kernel size, $n_e$ is the size of the latent channel assigned beforehand, and $n_l$ is the length of the embedding, in this case, $n_l = n_{cp}-8$.}\label{fig:network_structure}
	\end{subfigure}
	\caption{Illustration of the intention prediction network.}\label{fig:intention_prediction}
	\vspace{-0.4cm}
\end{figure}

\begin{figure}[t]
	\centering
	\begin{subfigure}{0.24\textwidth}
	    \centering
		\includegraphics[width =\textwidth]{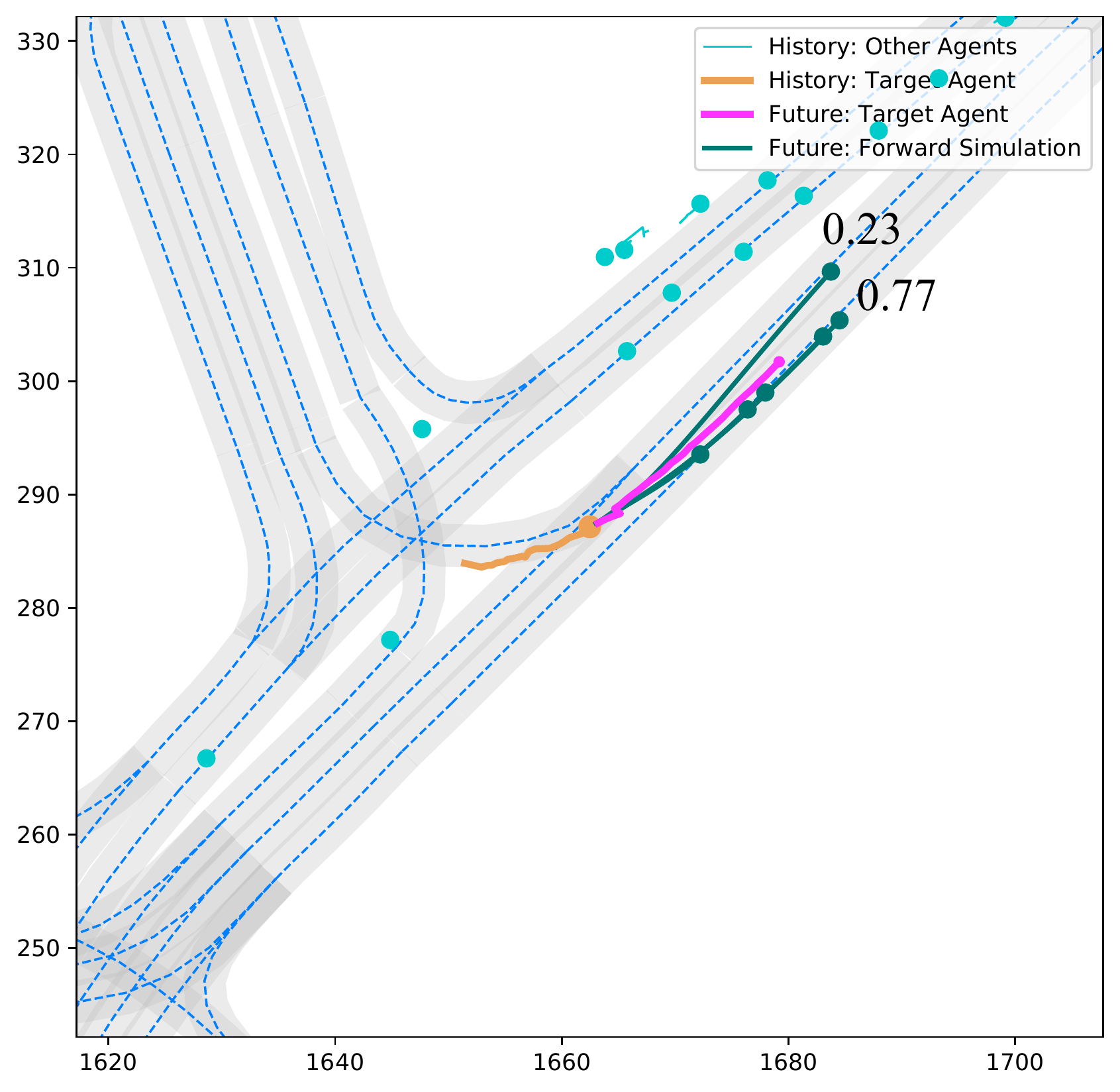}
	\end{subfigure}
	\begin{subfigure}{0.24\textwidth}
		\includegraphics[width =\textwidth]{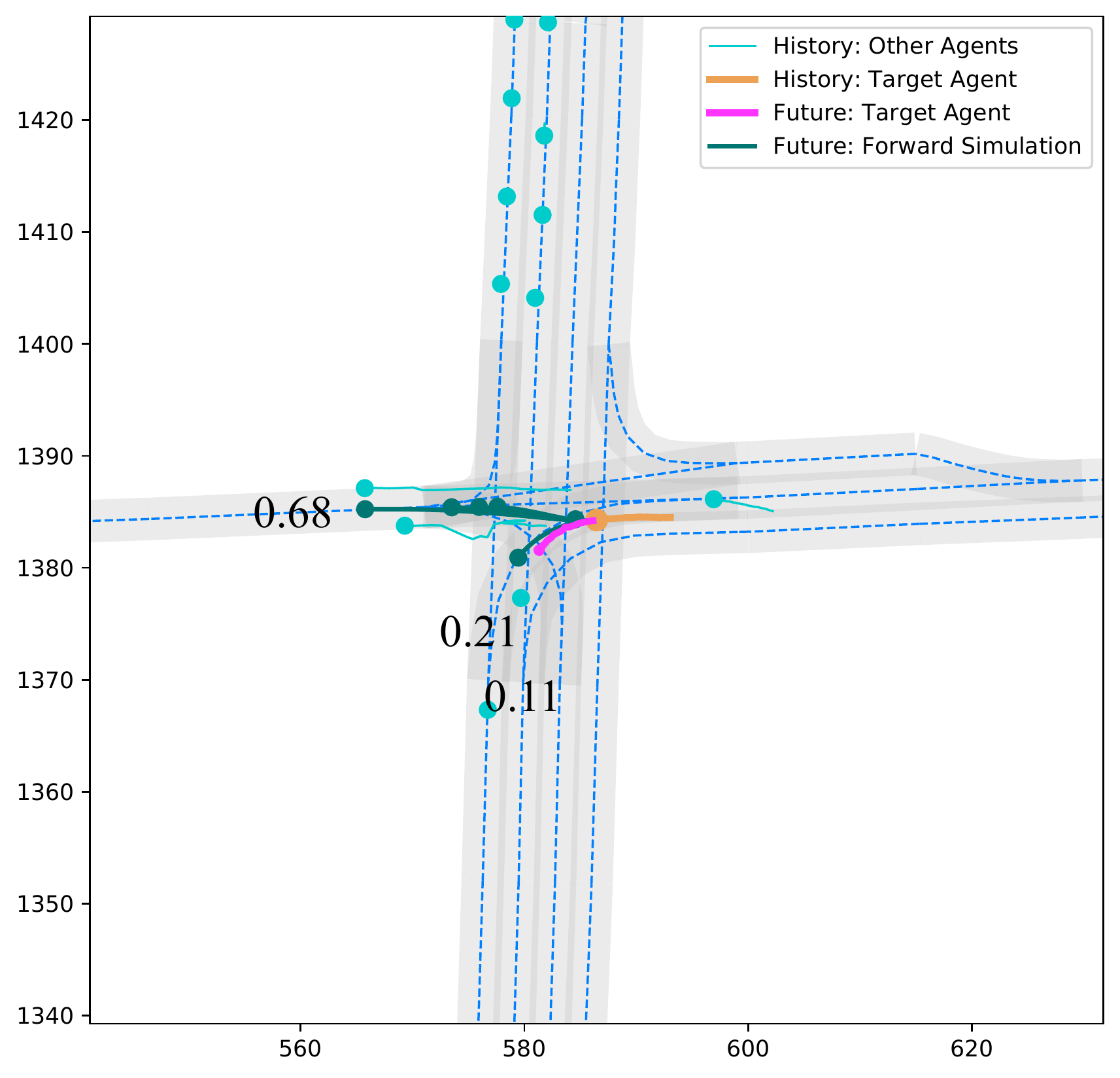}
	\end{subfigure}

	\begin{subfigure}{0.24\textwidth}
	    \centering
		\includegraphics[width =\textwidth]{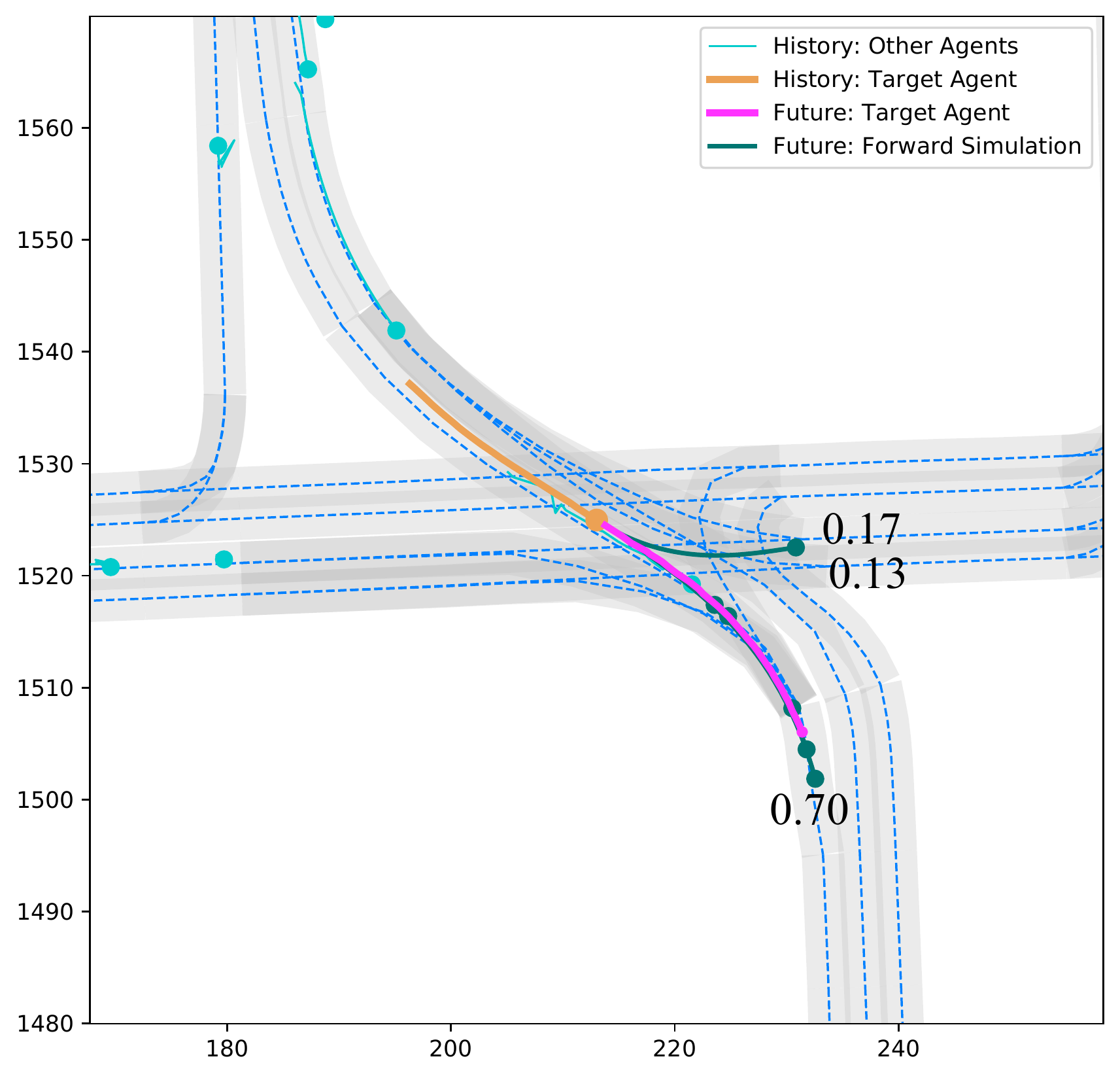}
	\end{subfigure}
	\begin{subfigure}{0.24\textwidth}
		\includegraphics[width =\textwidth]{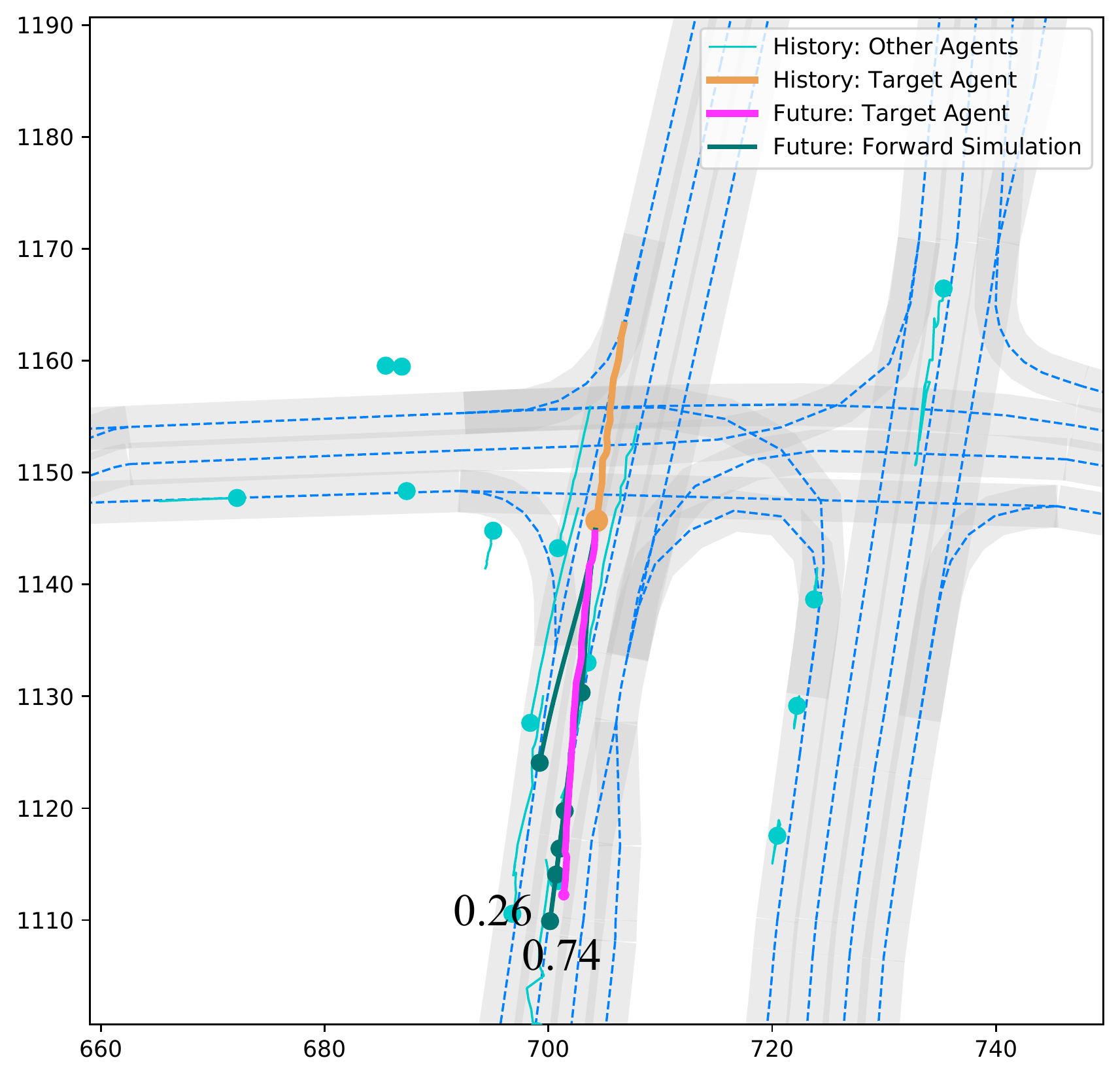}
	\end{subfigure}

	\caption{Intention estimation and forward simulation on Argoverse validation set.\label{fig:experiment_argoverse_validation}}
	\vspace{-1.0cm}
\end{figure}

To obtain better prediction performance, we present a neural network-based method to estimate the intention of the surrounding agents using the observation history and environmental information. The structure of the network is depicted in Fig.~\ref{fig:network}. At a high level, the network takes position samples from candidate center-lines and the history trajectory of the target vehicle as input features and uses an encoder-decoder structure to predict a scalar score for each candidate center-line. The score is then normalized by applying a softmax function. Since the network predicts the scalar score for each center-line instance and is not dependent on the number of center-lines $n_c$ in the scene, the network can work in various urban environments, as shown in Fig.~\ref{fig:experiment_argoverse_validation}.

As shown in Fig.~\ref{fig:intention_prediction}, three major modules are incorporated into the intention prediction network. The lane encoder takes the center-lines of the surrounding lanes of the target vehicle as input. We take the sampled points on the center-lines and use the position coordinates as input. For each center-line, we sample $n_{cp}=70$ points in total. The feature of the center-line is firstly encoded using multiple 1-D convolutional layers with kernel size equal to 3. Then we employ an average pooling in the tail. For the trajectory of the target vehicle, we use an LSTM to encode the 2-D positions recurrently and produce a latent vector, which is a compact representation of the motion history. We set the input size of the LSTM (after input embedding) to $n_x=4$ and the length of the hidden state in the LSTM to $n_h=64$. 
After processing the lane and trajectory information, we concatenate the trajectory embedding with every lane embeddings, then use a multilayer perceptron (MLP) with a ReLU activation function to estimate the score of the candidate lane. The final probability distribution is obtained via the softmax function. Fig.~\ref{fig:network_structure} shows the structure of the modules introduced above. Since the intention prediction is essentially a classification problem, we use the negative log-likelihood (NLL) as the loss to minimize.

\subsection{Policy Evaluation}\label{sec:policy_eval}
The overall reward for a policy is calculated by the weighted sum of the reward for each CFB-selected scenario. The reward function of each scenario consists of a linear combination of multiple user-defined metrics, efficiency cost $F_{e}$, safety cost $F_{s}$ and navigation cost $F_n$:
\begin{equation*}
    F_{total} = \lambda_1F_{e} + \lambda_2F_{s} + \lambda_3F_{n},
\end{equation*}
where $\lambda_1$, $\lambda_2$ and $\lambda_3$ are weights to balance the cost terms, and $F_{total}$ is the total cost (negative reward, i.e., $F_{total} = -R$).

To evaluate the efficiency cost, two sources are considered after a semantic action has finished. One item is the velocity difference between the current simulated velocity and the preferred velocity for the ego vehicle $\Delta v_{p}=|v_{{ego}}-v_{{pref}}|$. The other item partially represents the efficiency of the target lane by checking the velocity overshoot w.r.t. the leading vehicle $\Delta v_{o}=\max(v_{{ego}}-v_{{lead}}, 0)$, and calculating the velocity difference between the leading vehicle and the ego's preferred velocity $\Delta v_{l}=|v_{{lead}}-v_{{pref}}|$. Therefore, the efficiency of the policy can be written as
\begin{equation*}
    F_e = \sum^{N_a}_{i=0}F^i_e = \sum^{N_a}_{i=0}(\lambda^{p}_{e}\Delta v_{p} + \lambda^{o}_{e}\Delta v_{o} + \lambda^{l}_{e}\Delta v_{l}),
\end{equation*}
where $\lambda^{p}_{e}$, $\lambda^{o}_{e}$ and $\lambda^{l}_{e}$ are weights and $N_a$ is the number of semantic actions in a policy.

For the safety cost, we evaluate all the discretized states of the ego-simulated trajectory together with all other trajectories nearby. If a collision occurs, the corresponding scenario will be directly marked as failed and punished with a great cost. In addition, we also check whether the ego trajectory contains an RSS-dangerous state. If the state is RSS-dangerous, we obtain the safe velocity interval $\left[v^{lb}_{rss}, v^{ub}_{rss}\right]$ given the current situation, and punish the state by the difference from the safe velocity interval. The safety cost is written as
\begin{align*}
    F_s &= \lambda^{c}_{s}b_{\text{c}} + \sum^{N_s}_{i=0}b_{\text{r}}F^i_s\\
        &= \lambda^{c}_{s}b_{\text{c}} + \sum^{N_s}_{i=0}b_{\text{r}}v_{ego}\lambda^{r1}_{s}e^{\lambda^{r2}_s\lvert v_{ego} - \min(\max(v_{ego}, v^{lb}_{rss}),v^{ub}_{rss})\rvert},
\end{align*}
where $b_{\text{c}}$ and $b_{\text{r}}$ are the boolean values denoting the collision and the violation of RSS-safety for the ego simulated state, respectively. $\lambda^{c}_{s}$ is the penalty for collision. $N_s$ is the total number of the simulated states in the ego trajectory. The cost for RSS-dangerous is designed in an exponential form to emphasize the degree of the RSS-safety violation while $\lambda^{r1}_s$ and $\lambda^{r2}_s$ are tunable parameters.

The navigation cost is determined by the user preference and decision context. We enforce a reward (negative cost) for the policies that match the user's command provided by the human-machine interface (HMI), which can realize driver-triggered lane-change. Moreover, to improve the decision consistency, we also reward the policies with a similar meaning to the decision made by the last planning cycle. For example, if the optimal plan for the last cycle is ``changing to the left lane", we will assign a reward to the policies that achieve the same maneuver. Therefore, the cost term is represented as
\begin{equation*}
    F_n = \lambda^{user}_n b_{navi} + \lambda^{consist} b_{consist},
\end{equation*}
where $b_{navi}$ and $b_{consist}$ denote the flags of the policy that matches the navigation goal and decision history, respectively.

Note that the safety cost $F_s$ and efficiency cost $F_e$ are generated w.r.t. each semantic-level action. A discount factor $\gamma$ is introduced to adjust the weight for rewards in the distant future, and we set it to $\gamma=0.7$.

\section{Experimental Results}\label{sec:experimental_results}
In this section, we present the quantitative and qualitative analysis. The quantitative validations are based on the dataset and simulation, while the qualitative experiments are conducted on a real vehicle in dense real-world traffic.

\subsection{Quantitative Analysis}
We verify the effectiveness of the key features of EPSILON, namely, semantic-level action, interaction-awareness and the safety mechanism. We begin with a quantitative analysis on semantic-level action and forward simulation, since they determine how each scenario is realized. To verify the contributions of interaction-awareness and the safety mechanism, we conduct ablative studies for each component. 

\subsubsection{Semantic-level action and forward simulation}
As introduced in Section~\ref{sec:bp_forward_integration}, the semantic-level action and forward simulation are highly related to behavior/trajectory prediction techniques in the literature. The difference is that traditional prediction is often independent of planning, while our semantic-level action and forward simulation are coupled inside our behavior planner.

The reason why the semantic-level action and forward simulation can be coupled into planning is due to the simplification induced by the semantic-level actions. Specifically, our behavior planning reasons about a limited set of predefined closed-loop policies which allows for efficient and tight integration. However, it remains questionable whether this setup over-simplifies the problem and results in significant loss in the fidelity of future anticipation. To answer this question, we compare our intention estimation and forward simulation pipeline with two state-of-the-art learning-based trajectory prediction techniques, namely, TPNet~\cite{fang2020tpnet} and UST~\cite{he2020ust}. Both methods~\cite{fang2020tpnet, he2020ust} can generate multi-modal trajectory predictions with state-of-the-art accuracy.

\textbf{Dataset} We adopt the Argoverse~\cite{chang2019argoverse} motion forecasting dataset for comparisons, which is a large-scale autonomous driving dataset with rich map information. It provides a semantic graph which supports lane-level query. The dataset is collected in two different cities, Pittsburgh and Miami, including 205,942 sequences for training, 39,472 sequences for validation, and 78,143 sequences for testing. These three subsets are split with no geographical overlap.
Both TPNet~\cite{fang2020tpnet} and UST~\cite{he2020ust} have been also validated on this dataset.
We use 2-second past observations and predict spatial locations for the target vehicle in the future 3 seconds, which is the common setup used in~\cite{chang2019argoverse,fang2020tpnet, he2020ust}.

\textbf{Training} We implement the network using Pytorch~\cite{paszke2019pytorch}, and train the network on a single NVIDIA GTX 1080 GPU. The network is trained in an end-to-end fashion using Adam optimizer~\cite{kingma2014adam} with a batch size of 128. We initiate the learning rate to 1e-3 and decay to 1e-4 after 35 epochs. The total time cost is around 3 hours for 50 epochs.

\textbf{Metrics} We follow the official metrics of Argoverse to benchmark multiple predictions on this dataset. Average displacement error
(ADE) and final displacement error (FDE) are the most
used metrics in motion prediction. In order to evaluate the capability for modeling multi-modality, Argoverse~\cite{chang2019argoverse} also uses Minimum over N (MoN) metrics such as minADE and minFDE. Specifically, minFDE represents the distance between the endpoint of the best forecasted trajectory (i.e., the trajectory with the minimum endpoint error) and the ground truth, while minADE represents the average distance between the best forecasted trajectory and the ground truth. Following the setup of TPNet~\cite{fang2020tpnet} and UST~\cite{he2020ust}, we evaluate minADE and minFDE with six multi-modal trajectory predictions.

\begin{table}[t]
	\centering
	\caption{Comparison with baseline methods on Argoverse test set.}
	\begin{tabular}{@{}lcccc@{}}
	\toprule
	 \textbf{Methods}
	&\makecell{\textbf{ADE} \\ (m)}
	&\makecell{\textbf{FDE} \makeatletter @ \makeatother 3s \\(m)}
	&\makecell{\textbf{minADE}\\(m), N=6}
	&\makecell{\textbf{minFDE} \\(m), N=6} \\
	\midrule
	NN~\cite{chang2019argoverse} & 3.45 & 7.88 & 1.71  &  3.29\\
	LSTM ED~\cite{chang2019argoverse} & 2.96 & 6.81   & 2.34 &  5.44\\
	TPNet~\cite{fang2020tpnet} & 2.33 & 5.29 & 2.08  & 4.69\\
  TPNet-map~\cite{fang2020tpnet} & 2.23 & 4.71 & 2.04 & 4.23 \\
	TPNet-map-safe~\cite{fang2020tpnet}& 2.23 & 4.70 & 2.03 & 4.22\\
	TPNet-map-mm~\cite{fang2020tpnet}  & \textbf{2.23} & \textbf{4.70} & 1.61  & 3.28\\
	UST~\cite{he2020ust} &  - & - & 1.47  & 2.94\\
	\hline
	Our method & 2.58 & 5.70 & \textbf{1.41} & \textbf{2.48} \\
	\bottomrule
	\end{tabular}\label{tab:prediction_argoverse}
\end{table}

\begin{table}[t]
	\centering
	\caption{Ablative study on the impact of intention estimation accuracy on Argoverse validation set.}\label{tab:}
	\begin{tabular}{@{}lcccc@{}}
	\toprule
	\textbf{Methods}
	&\makecell{\textbf{ADE}\\m}
	&\makecell{\textbf{FDE} \\\makeatletter @ \makeatother 3s}
	&\makecell{\textbf{minADE}\\(m), N=6}
	&\makecell{\textbf{minFDE} \\(m), N=6} \\
	\midrule
	Without intention estimation & 3.94 & 9.13 & 2.07  &  4.16\\
	With intention estimation & 2.19 & 4.80   & 1.20 &  2.05\\
	Oracle intention estimation & 2.12 & 4.58 & 1.13  & 1.83\\
	\bottomrule
	\end{tabular}\label{tab:ablative_argoverse}
\end{table}

\textbf{Baselines} All the baseline methods are compared on the Argoverse test set. Depending on whether map information or prior knowledge is used, TPNet~\cite{fang2020tpnet} has several variants.
Apart from TPNet~\cite{fang2020tpnet} and UST~\cite{he2020ust}, we also include another two baselines, namely, NN and LSTM ED, which are commonly used in comparisons.

\begin{itemize}
	\item Nearest Neighbor (NN)~\cite{chang2019argoverse}: weighted nearest neighbor regression using top-N (N=6) hypothesized center-lines.
	\item LSTM ED~\cite{chang2019argoverse}: LSTM Encoder-Decoder network with road map information as input.
	\item TPNet~\cite{fang2020tpnet}: TPNet with only 2-second past observations and without road map information.
	\item TPNet-map: TPNet with road map information as input.
	\item TPNet-map-safe: TPNet with prior knowledge for constraining proposals.
	\item TPNet-map-mm: TPNet with prior knowledge for generating multiple proposals.
	\item UST~\cite{he2020ust}: Unified spatio-temporal representation for learning-based trajectory prediction.
\end{itemize}

\textbf{Results} As listed in Tab.~\ref{tab:prediction_argoverse}, for prediction accuracy with only one trajectory hypothesis, our method achieves $2.58$ for ADE and $5.7$ for FDE, outperforming NN and LSTM ED baselines, but there is still a gap w.r.t. to the two state-of-the-art baselines. When considering multiple prediction hypotheses, which is the common case in real-world driving, our method outperforms all the baseline methods (incl. the two state-of-the-art methods) significantly, achieving $1.41$ for minADE and $2.48$ for minFDE.
Note that the motivation for introducing the semantic-level action and forward simulation is not targeting ``prediction accuracy''.
Instead, it aims at providing a lightweight querying interface which can be coupled inside planning for realizing diverse future scenarios conditioned on the ego decision.
To our surprise, despite our lightweight design, our method still achieves satisfactory accuracy, which supports our claim that simplified predefined controllers are sufficient for faithfully capturing the future scenarios.

To further study the impact of the accuracy of intention estimation on the scenario realization, we conduct an ablative study in which we replace the intention estimation network with a dummy predictor which generates a uniform probability distribution on all candidate center-lines. On the other hand, we set up an oracle intention estimator which uses the future ground truth center-line for the forward simulation. We compare these two alternatives with our method on the Argoverse validation set, as shown in Tab.~\ref{tab:ablative_argoverse}. We find that without an intention estimation module, the accuracy of forward simulation drops significantly in terms of ADE and FDE, and also in terms of minADE and minFDE, which indicates that it is important to include an intention estimation module (even a lightweight one) in complex environments. On the other hand, with an oracle intention predictor, the accuracy is improved but the gap is relatively small. Our conjecture is that since the forward simulation is based on predefined controllers, when the intention estimation becomes perfectly accurate, the accuracy bottleneck is on the predefined models. In the future, we may explore using learning-based controllers for the forward simulation, as we previously attempted in~\cite{song20pip}. However, we think only marginal improvements can be made, considering that the accuracy of the existing method (a lightweight intention estimator with a model-based forward simulator) is already high (since it outperforms the two state-of-the-art methods).

\subsubsection{Impact of coupling prediction and planning}
\begin{figure}[t]
	\centering
	\includegraphics[width=0.4\textwidth]{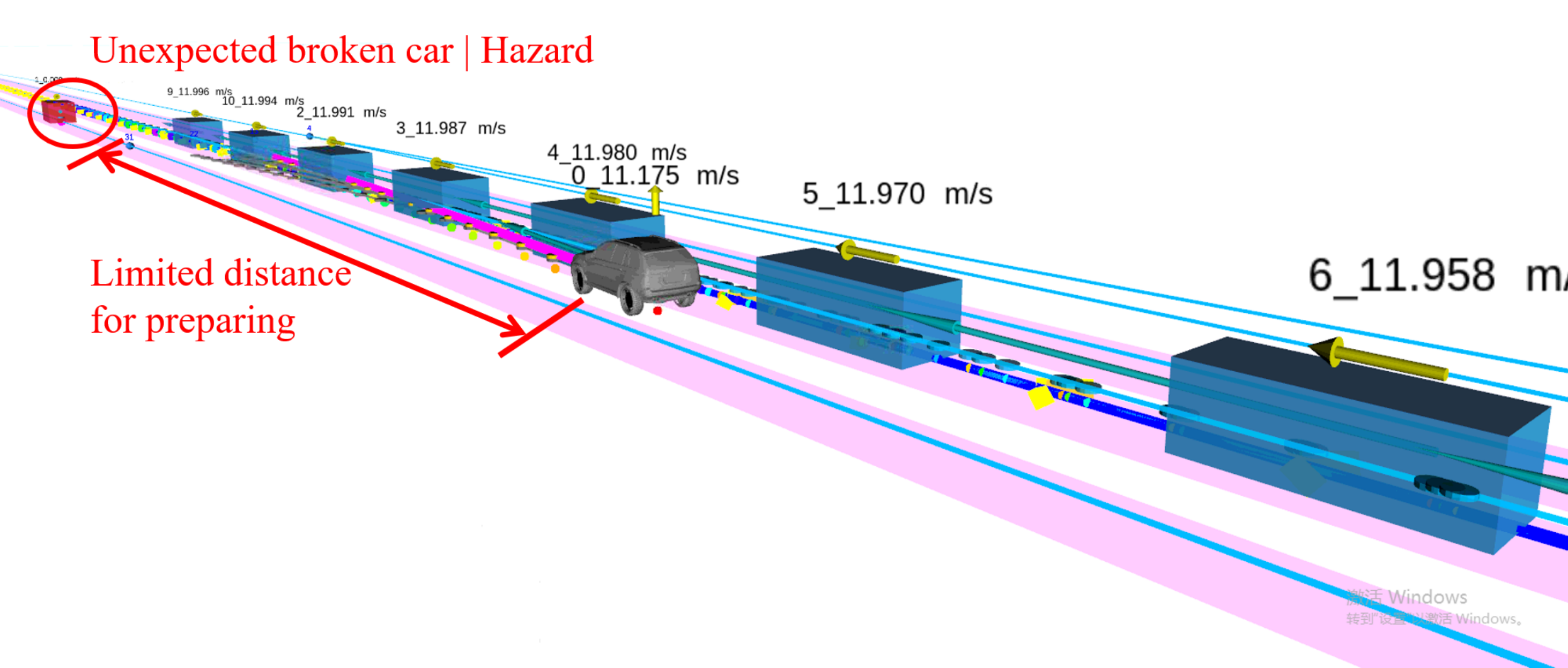}
	\caption{Illustration of the benchmark scenario.}\label{fig:benchmark_scenario}
\end{figure}

To quantitatively examine the impact of coupling prediction and planning, we present a benchmark scenario for testing. The setup of the benchmark scenario follows a typical driving configuration which we often encounter in city driving, as shown in Fig.~\ref{fig:benchmark_scenario}. In this scenario, there is an unexpected broken-down car ahead, and the controlled vehicle should react in a limited preparation distance. If the controlled vehicle fails to grasp the opportunity to change lanes, it is forced to slow down, which not only affects its travel efficiency but also leads it into a hazard as it needs to wait indefinitely for the incoming traffic to clear. However, changing lanes may also be risky. On the one hand, the controlled vehicle should not accelerate too much, considering the risk of collision in the case of sudden braking by the leading vehicle in the nearby lane, while on the other, it should complete the lane change in a timely way to be safe (in the sense of responsibility) from a rear collision with the following vehicle. To achieve good performance in this benchmark, the planner is required to conduct counterfactual reasoning, namely, examining how different future decisions of the ego vehicle affect its surrounding traffic participants. 

Note that in the simulation, all other traffic participants (AI agents for short) are controlled by their own policy according to the perception information they received. The policy for AI agents is a simple adaptive cruising policy adapted from the intelligent driver model~\cite{treiber2013traffic} that mimics a real traffic queue. We define three aggressiveness levels by adjusting the \textit{desired time headway} that directly determines the longitudinal distance to the leading vehicle. We also define the cooperative range to describe the friendliness of the AI agents to the queue-jumper vehicle, namely, the AI agents will yield and keep distance to the cut-in vehicle when the lateral distance between the vehicle centroid and the current center-line is smaller than the cooperative range. To make a fair experiment, the settings such as the aggressiveness level and cooperative range of the AI agents is unknown to the ego planning system.

\begin{table}[t]
	\centering
	\caption{Comparison with decoupled prediction and planning system and EPSILON w/o safety mechanism. The time in the brackets denotes the value of \textit{desired time headway}.}
	\begin{tabular}{@{}lcccc@{}}
	\toprule
	 \textbf{Methods}
	&\textbf{\makecell{Aggress.\\level}}
	&\textbf{\makecell{Cooperative\\range (m)}}
	&\textbf{\makecell{Travel effi. \\ ($m/s$)}}
	&\textbf{\makecell{Safety\\cost}}\\
	\midrule
	Decoupled & 1 (2.0 $s$) & 2.55 & 11.1 (\cmark)   & 4.5\\
	EPSILON w/o safety  & 1 (2.0 $s$) & 2.55 & 11.1 (\cmark) &  3.85 \\
	EPSILON & 1 (2.0 $s$) & 2.55   & 11.1 (\cmark) &  3.05\\
	\hline
	Decoupled & 2 (1.5 $s$) & 2.00 & 10.6 (\cmark) &  81.4\\
	EPSILON w/o safety  & 2 (1.5 $s$) & 2.00 & 10.5 (\cmark) &  65.9 \\
	EPSILON  & 2 (1.5 $s$) & 2.00 & 10.6 (\cmark) & 13.7 \\
	\hline
	Decoupled & 3 (1.0 $s$) & 1.75 & 7.3 (\xmark)  &  42.3\\
	EPSILON w/o safety  & 3 (1.0 $s$) & 1.75 & 10.7 (\cmark) &  147.2\\
	EPSILON  & 3 (1.0 $s$) & 1.75  & 10.8 (\cmark) &  16.8\\
	\bottomrule
	\end{tabular}\label{tab:comparison_benchmark}
	\vspace{-1.0cm}
\end{table}

\textbf{Metrics} In the benchmark scenario, \textit{safety} and~\textit{travel efficiency} are two representative metrics. Successful merging leads to higher travel efficiency, since failure of lane change results in waiting indefinitely in the hazard. However, higher efficiency should not come at the price of sacrificing safety. The travel efficiency metric can be simply represented by the average velocity, while the safety metric can be represented by RSS. Specifically, we evaluate the safety cost of the chosen decision (i.e., a sequence of states) for every planning cycle and calculate the average. We quantitatively evaluate the average travel efficiency of the trace of the ego vehicle and the average safety cost of each decision for the first $300$ planning cycles (around 15 seconds) which are the time needed to safely pass the hazard. We vary the aggressiveness and cooperative range of the traffic participants in the nearby lane to adjust the level of difficulty.

\begin{figure}[t]
	\centering
	\begin{subfigure}{0.4\textwidth}
	    \centering
		\includegraphics[width=\textwidth]{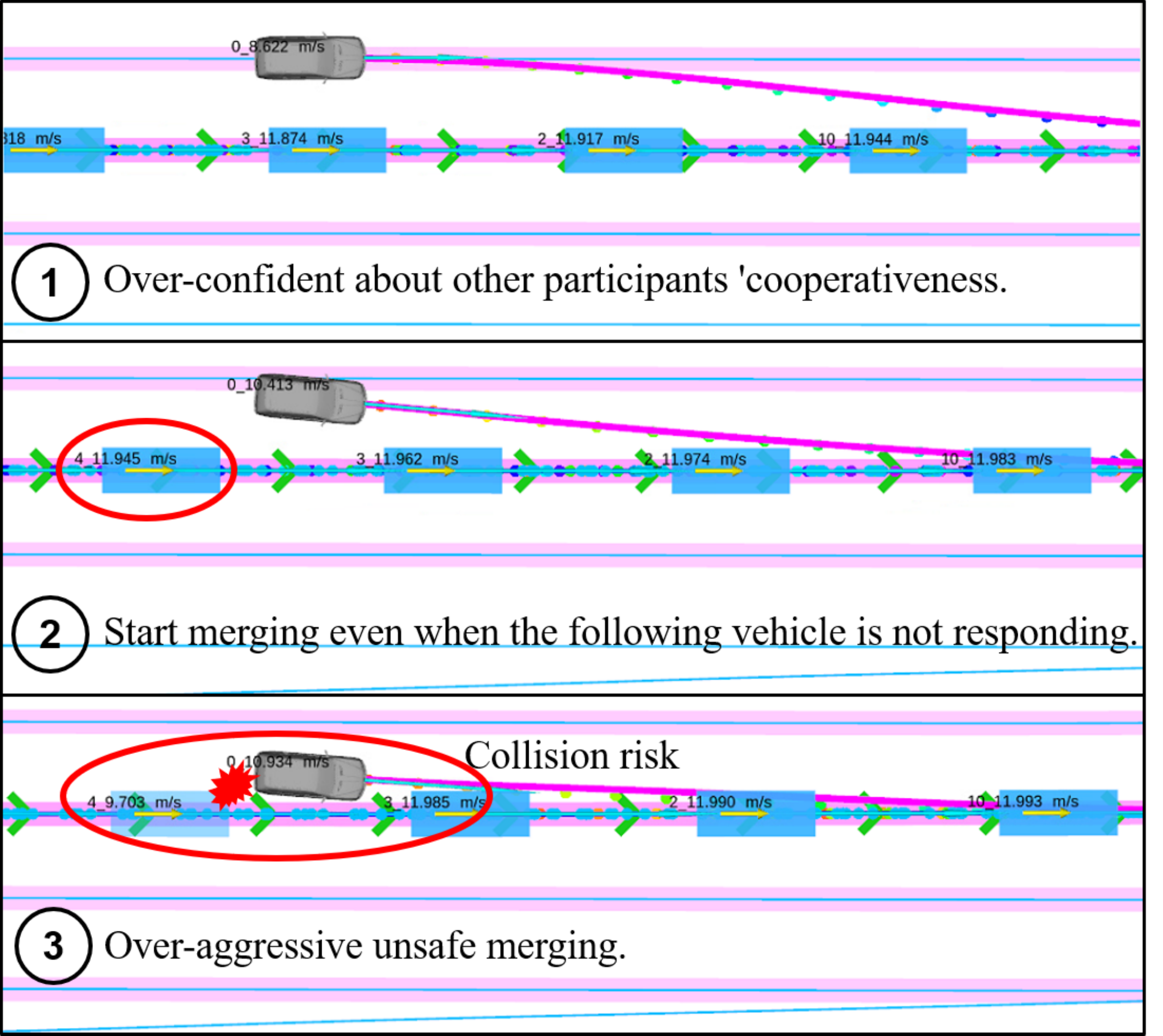}
		\caption{Merging w/o safety mechanism}\label{fig:benchamrk_merging_unsafe}
	\end{subfigure}
	
	\begin{subfigure}{0.4\textwidth}
		\includegraphics[width=0.99\textwidth]{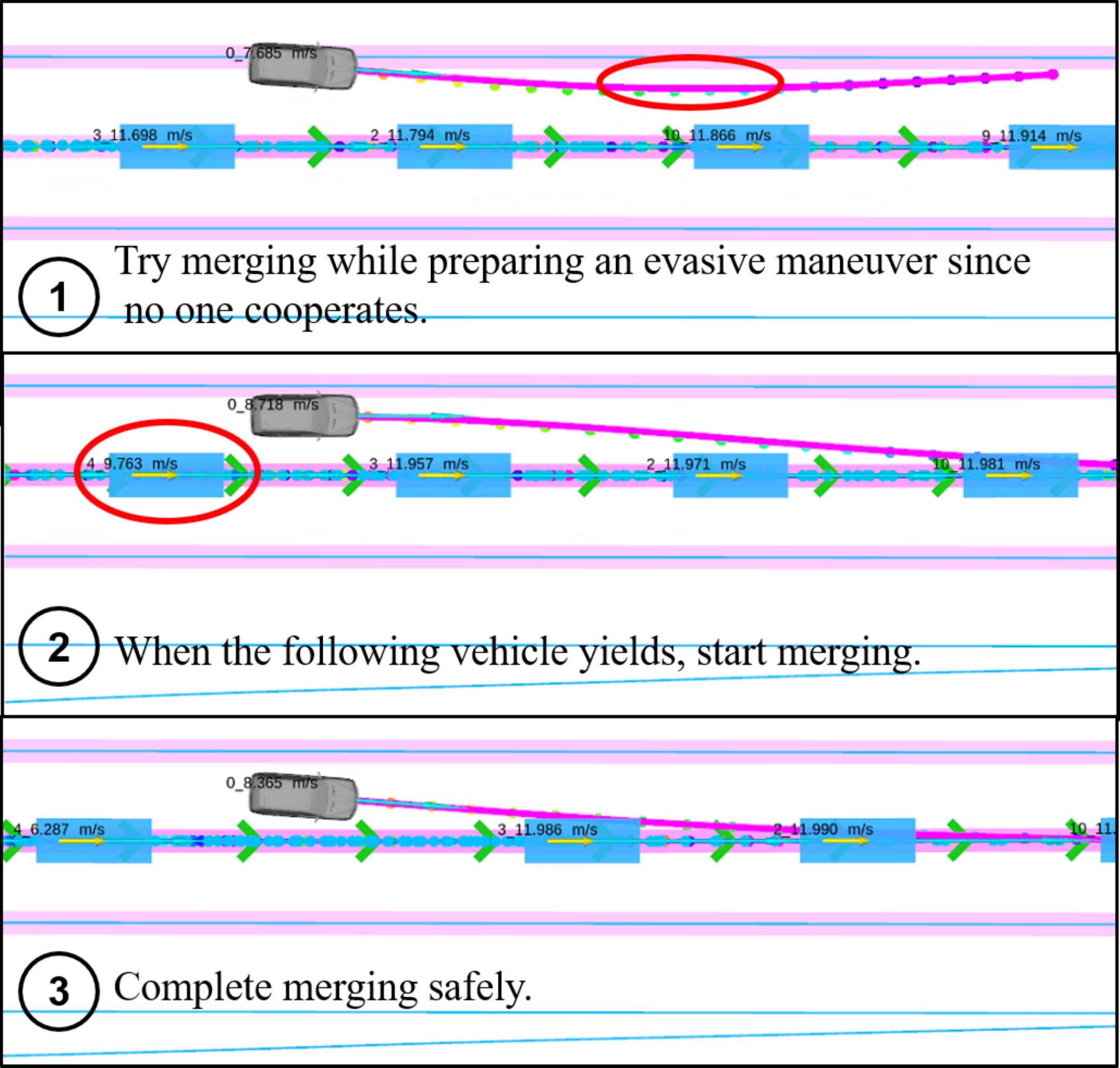}
		\caption{Merging w/ safety mechanism}\label{fig:benchamrk_merging_safe}
	\end{subfigure}
	\caption{Illustration of merging into dense traffic w/ and w/o safety mechanism. Note that the decoupled prediction-planning method is excluded here since it fails to conduct the lane change and gets trapped in the hazard. The best decision is visualized by the~\textit{rainbow} dots, while the corresponding trajectory is marked in~\textit{purple}.}\label{fig:benchamrk_merging}
	\vspace{-0.5cm}
\end{figure}

\textbf{Baseline} The baseline for the comparison is the traditional decoupled prediction and planning pipeline. Specifically, we use an independent trajectory prediction module which is derived from multi-agent forward simulation, but exclude the decision of the controlled vehicle. The resulting prediction is unaware of the impact of the ego's future action, while retaining the interaction among other traffic participants, which maintains the exact functionality of the popular trajectory prediction methods~\cite{fang2020tpnet,he2020ust}. After obtaining the predicted trajectory, we generate multiple action sequences in the same way as EPSILON and pick out the best action sequence using the same evaluation pipeline as for EPSILON.

\textbf{Results} As shown in Tab.~\ref{tab:comparison_benchmark}, we find that, in less aggressive and cooperative traffic (aggressiveness 1 with $2.0$ s headway time and 2.55 $m$ cooperative range), the performance of the decoupled planning system is close to that of EPSILON. Both methods can conduct lane change successfully with high travel efficiency while also staying safe. However, when increasing the aggressiveness of other traffic participants to $3$ with $1.0$ s headway time, and narrowing down the cooperative range to $1.75$ m, decoupled prediction-planning method cannot find a gap in which to merge in the preparation range (marked by \xmark), resulting in waiting in the hazard. Note that the time of interest is set to the first $300$ planning cycles, and when considering a longer horizon for calculating statistics, the average velocity of the decoupled method can be even lower due to waiting for the clearance of all the traffic. By contrast, EPSILON can successfully achieve safe lane change with high travel efficiency ($10.8$ $m/s$) and a small safety cost $16.8$. The reason is that the ego vehicle can implicitly negotiate with the traffic participants in the nearby lane by considering the impact of different future decisions. To summarize, the performance gain from using the coupled prediction-planning framework appears in dense and aggressive traffic.

\subsubsection{Impact of the safety mechanism}
As introduced in Section~\ref{sec:safety_mechnism}, both our previous method~\cite{ding2020eudm} and MPDM~\cite{cunningham2015mpdm} suffer from the problem that the predefined policies may be over-confident, which results in over-aggressive maneuvers. To overcome this, we introduce a safety mechanism in Section~\ref{sec:safety_mechnism}. To verify its effectiveness, we set up a baseline which excludes the safety mechanism of EPSILON (``EPSILON w/o safety'' in Tab.~\ref{tab:comparison_benchmark}), which essentially represents the performance of our previous method~\cite{ding2020eudm}. The benchmark scenario and metrics are the same as those in the previous section.

As shown in Tab.~\ref{tab:comparison_benchmark}, we find that the performance gap between ``EPSILON w/o safety'' and EPSILON is quite small in loose and cooperative traffic. However, in aggressive and less cooperative traffic, for example, when the aggressiveness level is 3 with a 1.75 $m$ cooperative range, there is a huge gap in terms of the safety cost of resulting decisions. Although EPSILON w/o safety can conduct the lane change and achieve good travel efficiency with an average speed of $10.7$ $m/s$, it is over-confident about other traffic participants' cooperativeness, which incurs an average safety cost of $147.2$, which is $8.7$ times that of EPSILON.

A typical example is provided in Fig.~\ref{fig:benchamrk_merging} to illustrate how the difference arises. As shown in Fig.~\ref{fig:benchamrk_merging_unsafe}, EPSILON w/o safety always assumes the following vehicle in the nearby lane will yield according to the predefined model (e.g., IDM). However, when the following vehicle is not as cooperative as the planner pre-assumes, the controlled vehicle will pick out a decision which is over-aggressive. In this case, EPSILON w/o safety leads to a near-collision scenario. In contrast, thanks to the safety mechanism, when the controlled vehicle tries to change lanes, EPSILON incorporate the lane change with an backup maneuver. With this maneuver, the controlled vehicle does not need to ``estimate'' the cooperativeness of the following vehicle and can always assume it is not cooperative. However, this conservative assumption does not prevent the controlled vehicle from conducting the lane change due to the existence of the backup plan. The resulting situation becomes that the ego vehicle will first adjust itself to a position which is safe, and merges when the following vehicle responds. This ``push-wait-merge'' style of merging is automatically achieved by the safety mechanism instead of using complex handcrafted logics.

\subsubsection{Computation Time Cost}
\begin{figure}[!htb]
	\centering
	\includegraphics[width =0.38\textwidth]{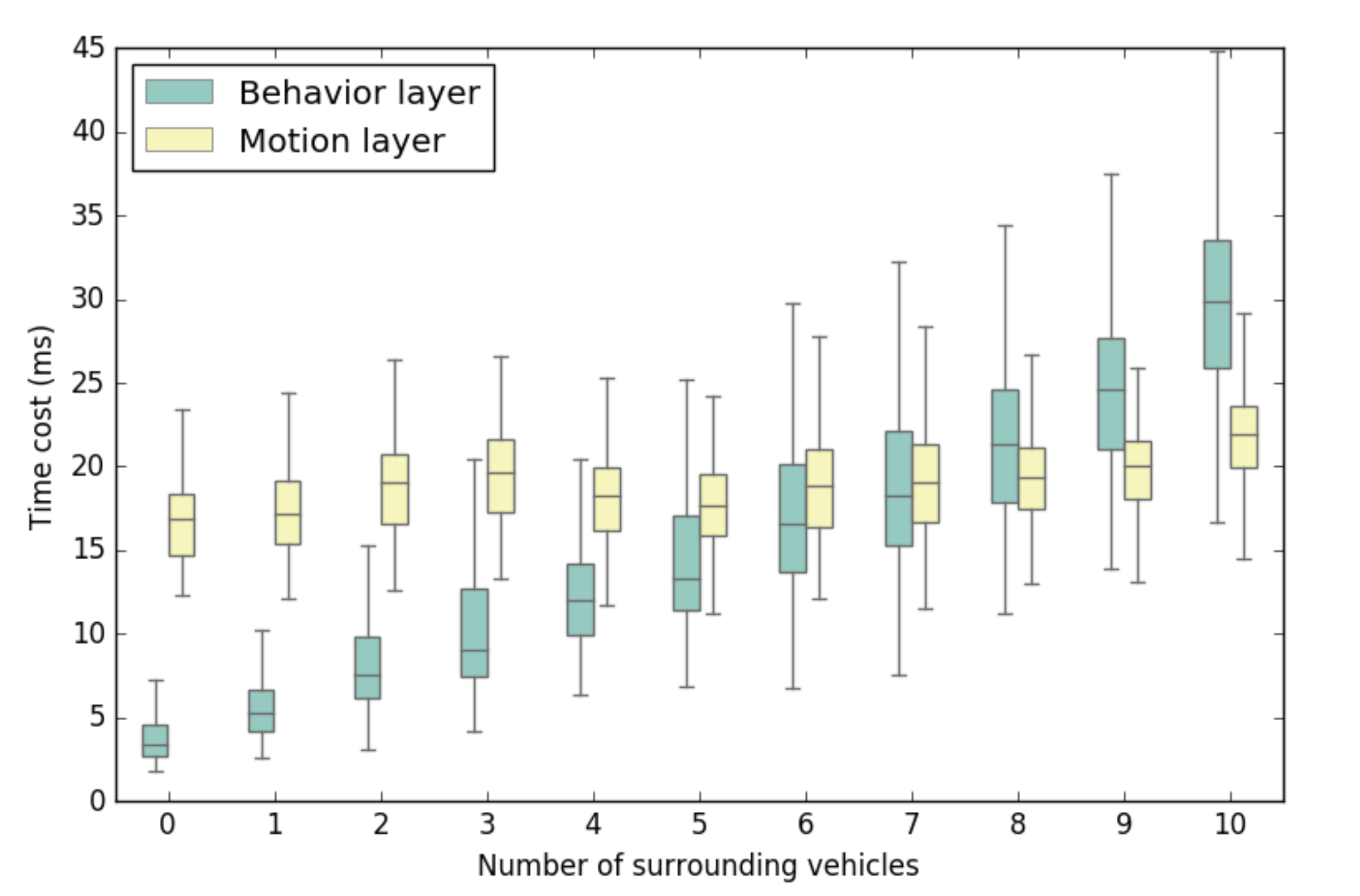}
	\caption{Illustration of the computational time required by the proposed planning system w.r.t. the number of vehicles considered. Results for both the behavior layer and motion layer are presented.}\label{fig:time_cost}
	\vspace{-0.4cm}
\end{figure}

To verify whether the proposed planning system meets the requirement of real-time applications, we analyze the computational time cost of both the behavior and motion layer in the simulation on a consumer-level desktop (Intel i7-8700 with 32GB RAM). By changing the number of vehicles considered, we can obtain the computational time cost from simple to complex scenarios, and for each situation, we collect over 250 frames for statistical analysis. As shown in Fig.\ref{fig:time_cost}, the time cost of the behavior planner increases w.r.t. the number of surrounding vehicles, while the time cost of the motion planner is basically constant. We can find that the time cost of these two modules is all restricted within 50 milliseconds, and by leveraging the pipeline structure mentioned in Sec.~\ref{sec:system_overview}, our EPSILON can run stably at 20 Hz.

\subsection{Qualitative Analysis}

In this section, we present the onboard experimental results in the real-world traffic. Our experiment platform is a prototype vehicle refitted from a passenger-SUV, which is equipped with a sensing suite including two sets of forward-facing stereo cameras, four fish-eye cameras, four millimeter-wave radars, six ultrasonic sensor and a low-cost INS. Neither Lidar nor high-precision positioning system is applied in this platform. Note that the HD-map is excluded during the real-world road tests, both localization information and road structure are inferred online. And the other traffic participants (surrounding vehicles) are detected and tracked by the perception module. Our planning system receives the data from these upper-level modules and sends the planned trajectory to the control system. The vehicle control system is based on the model predictive control (MPC) which achieves accurate tracking performance in both low and high-speed driving. For onboard testing, our planning system is implemented using C++ and deployed on a single NVIDIA Xavier with an 8-core ARM-base CPU.

\begin{figure}[t]
	\centering
	\begin{subfigure}{0.24\textwidth}
	    \centering
		\includegraphics[width = \textwidth]{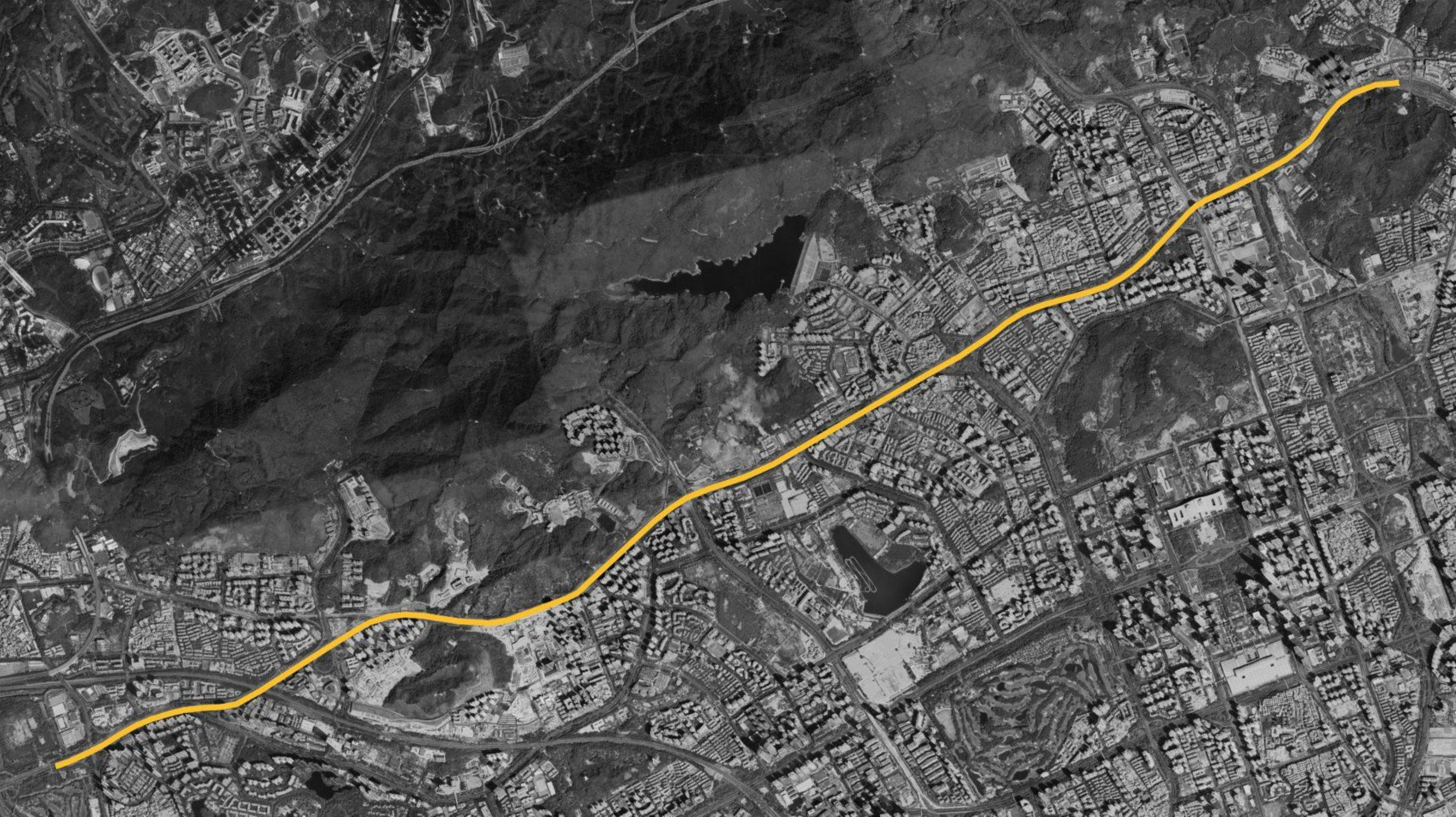}
		\caption{Test route is marked in yellow.}
	\end{subfigure}
	\begin{subfigure}{0.24\textwidth}
		\includegraphics[width = \textwidth]{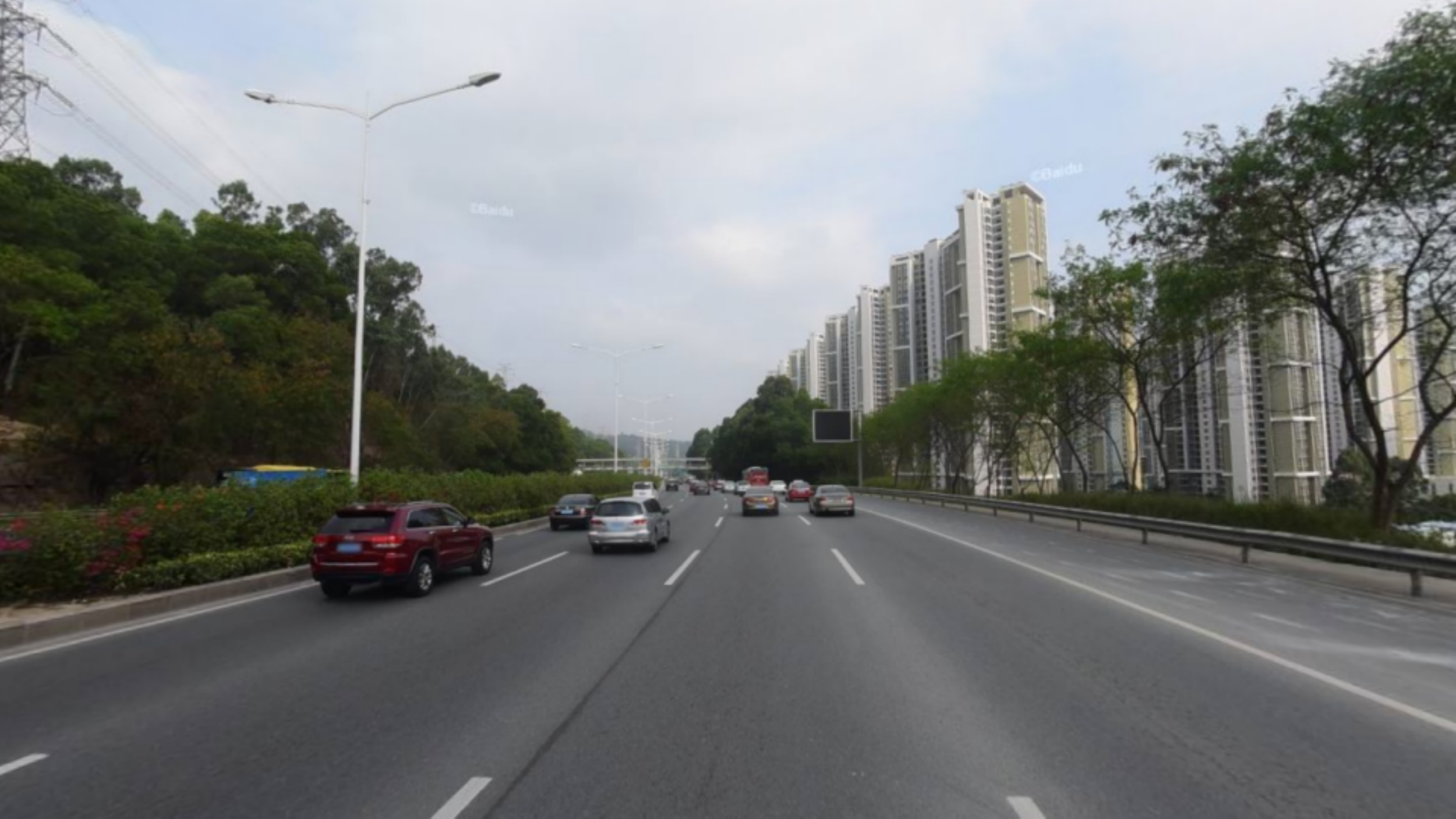}
		\caption{Street view of the test road.}
	\end{subfigure}
	\caption{Illustration of the test road. The length of the test route is approximately 12.5 kilometers and contains up to four lanes in each direction. Unlike other intercity expressway, there is an entrance or exit every 500 meters on average, so vehicle interactions on this road are very common. (Satellite imagery credit: Google.)}\label{fig:test_road}
	\vspace{-1.0cm}
\end{figure}

We validate our system in a busy urban expressway shown in Fig.~\ref{fig:test_road}. The test road is a four-lane dual carriageway with a speed limit of 80 km/h and the total length of the test route is about 12.5 kilometers. We conduct four test drives on this same route. For the first two trials, we disable the active lane-changing option and we encourage the human driver to trigger lane change command as he wants. For the other two trials, we enable the active lane change and try to rely only on the lane change decision proposed by the planner.

\begin{figure}[!htb]
	\centering
	\includegraphics[width =0.4\textwidth]{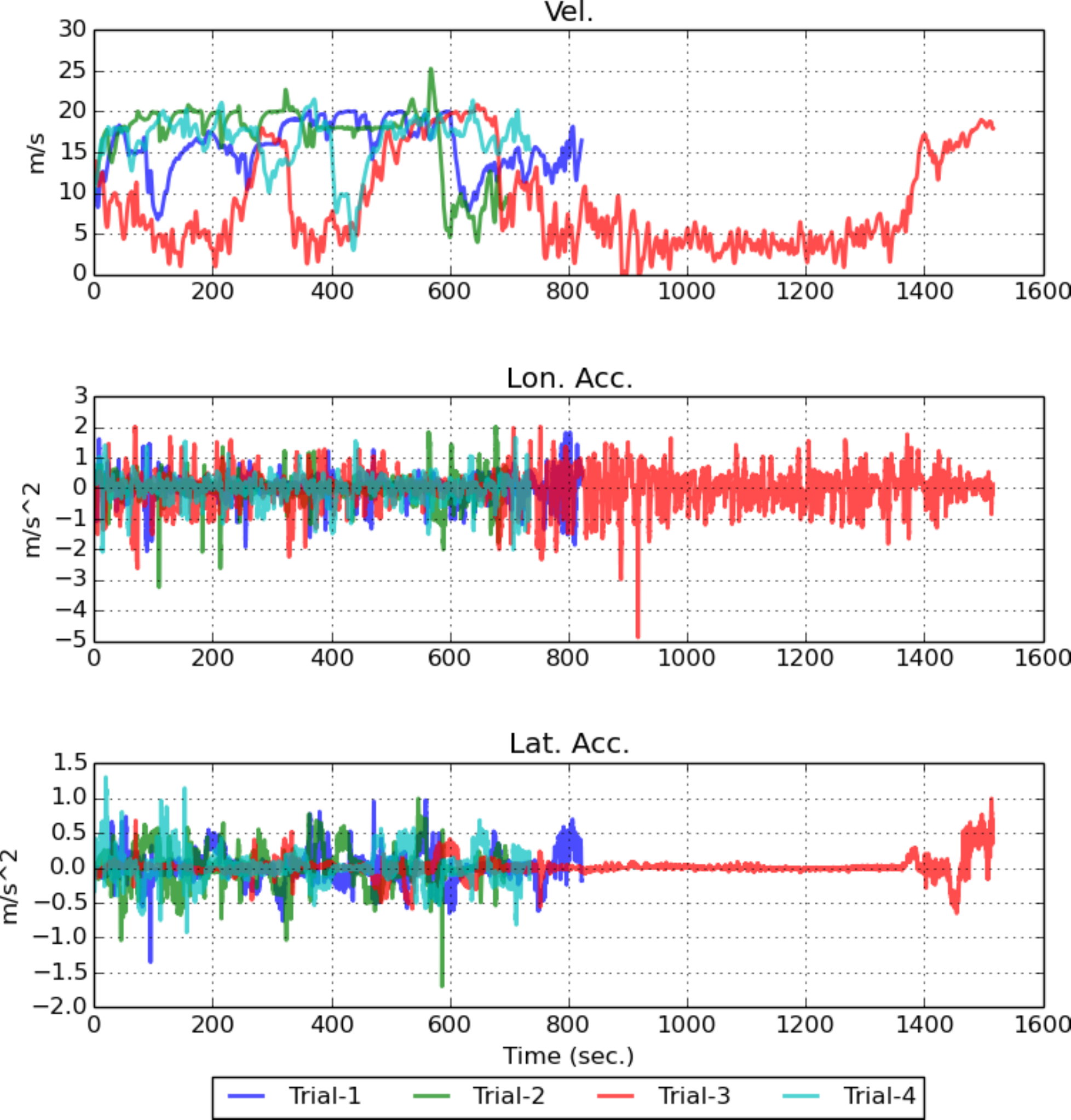}
	\caption{Dynamic profiles for the four test drives. The time duration of the third trial is much longer than the others since the road is quite congested during the experiment. With the exception of a few rare cases, the vehicle's dynamics were kept within a comfortable range. For Trial-2 at around 590-th second, a lane-cancel behavior was executed in a risky situation, causing a relatively large lateral acceleration up to 1.60$m/s^2$ (actually still acceptable). At around 920-th second of Trial-3, the controlled vehicle made a hard brake (-4.88$m/s^2$) because of a drastic deceleration of the leading vehicle. }\label{fig:long_term_curve}
\end{figure}

The statistics of the evaluation metrics are shown in Tab.~\ref{tab:metrics_test_drives}. As we can see, all the maximum longitudinal and lateral accelerations are limited in an appropriate range resulting in a comfortable ride (also see Fig.~\ref{fig:long_term_curve}). Due to the presence of many ramp entrances and exits, the surrounding traffic participants will frequently change lanes and merge in front of the ego vehicle, causing plenty of interactions. Among the four test drives, we achieve a reasonable overall success rate for lane-changing (over 80\%) in such highly interactive environments. Note that lane change cancellations handled by the backup plans are not recorded as success cases. Besides, we observe that human-like maneuvers such as ``find gap then merge" and cut-in handling can be achieved automatically. Despite risky scenarios are inevitable in such interactive driving environments, our planner can handle them appropriately by leveraging the proposed backup plans and reduce the number of human interventions.
We find that the driving time duration for the third trial is much longer than others because the test vehicle encounters traffic jams during the experiment, leading to a much lower average velocity and more challenging driving scenarios. As a result, the success rate of lane change drops a lot in the third trial (62.5\%), and the number of human intervention is larger (3 times in total). We blame the performance degeneration in two aspects: firstly, since the traffic jams often appear near the ramp exits and entrances, traffic participants are more likely to conduct aggressive maneuvers leading to more risky scenarios; secondly, in congested traffic, the noises of sensor data increase since there is more occlusion due to the complex environment, resulting in inaccurate evaluations and improper decisions. Besides, the parameters of our planner are tuned to be more proactive causing more aggressive maneuvers during the experiments. We find nearly half of the interventions are due to improper trigger time for active lane change under noisy observations (3 times), and the rest of them are due to 1) blocked by aggressive drivers (twice), 2) infeasible plan caused by the misestimation of others' intention (once), and 3) drastic deceleration of the leading vehicle (once). Here, we point out that simply reducing the number of human interventions is not a major target of this work since once we tune the planner into a conservative style, metrics such as miles per intervention (MPI) will increase significantly. At the same time, we expect the planner that produces as flexible behaviors as possible while preserving driving safety.

\begin{table}[t]
	\centering
	\caption{Evaluation metrics of the four test drives}
	\begin{tabular}{@{}cccccc@{}}
	\toprule
	&\textbf{\makecell{Trial 1}}
	&\textbf{\makecell{Trial 2}}
	&\textbf{\makecell{Trial 3}}
	&\textbf{\makecell{Trial 4}}
	&\textbf{Sum} \\
	\midrule
	Avg vel ($km/h$)        & 56.2        & 61.0        & 30.8        & 59.3        & - \\
	Min vel ($km/h$)        & 24.2        & 14.2        & 0.0         & 10.7        & - \\
    Max vel ($km/h$)        & 72.4        & 90.9        & 74.9        & 77.5        & - \\
    Max lon acc ($m/s^2$)   & 1.83        & 2.00        & 2.02        & 1.65        & - \\
    Max lon dec ($m/s^2$)   & -2.07       & -3.24       & -4.88       & -2.08       & - \\
    Max lat acc ($m/s^2$)   & 1.35        & 1.60        & 0.99        & 1.29        & - \\
    Time duration           & 14$'$42$''$ & 12$'$35$''$ & 25$'$16$''$ & 12$'$14$''$ & - \\ \hline
    Lane change triggered   & 14          & 17          & 8           & 10          & 49 \\
    Lane change success     & 12          & 15          & 5           & 8           & 40 \\
    Backup plan triggered   & 3           & 2           & 3           & 2           & 10 \\
    Backup plan success     & 2           & 1           & 1           & 0           & 4 \\ \hline
    \makecell{Human take-over for\\unsafe lane change}  & 1 & 1 & 2 & 2 & 6  \\
    \makecell{Human take-over for\\unsafe distance}     & 0 & 0 & 1 & 0 & 1  \\
	\bottomrule
	\end{tabular}\label{tab:metrics_test_drives}
	\vspace{-1.0cm}
\end{table}

To show the flexibility and robustness of our planning system in detail, we provide four representative cases collected from our daily road tests.

\subsubsection{Intelligent gap finding and merging}
\begin{figure*}[t]
	\centering
	\begin{subfigure}{0.9\textwidth}
	    \centering
		\includegraphics[width =\textwidth]{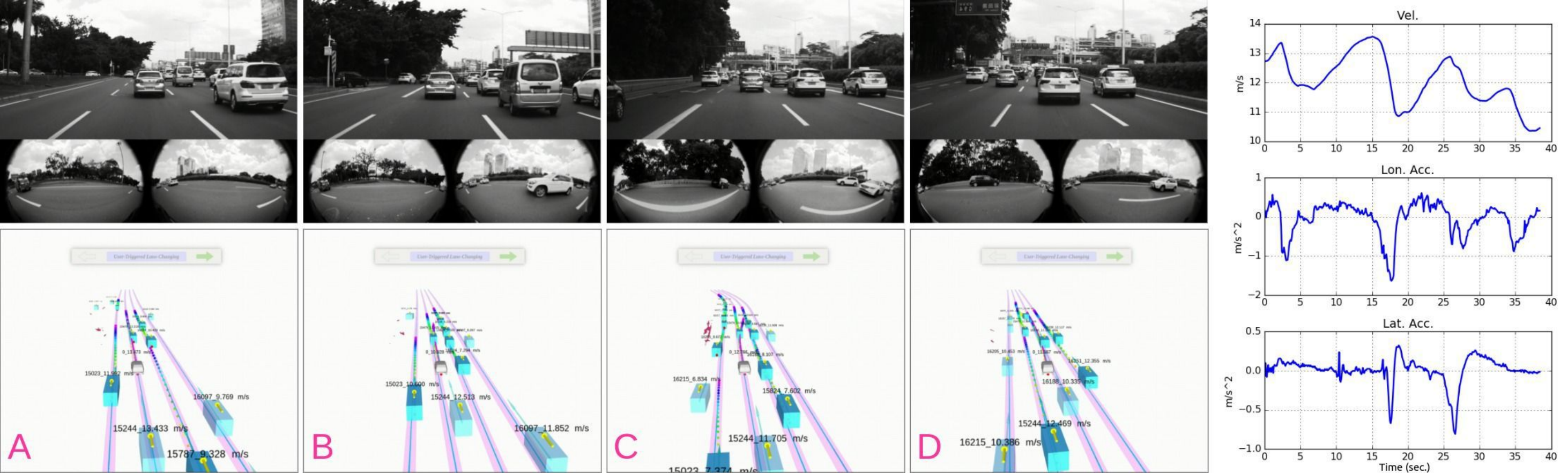}
		\caption{Automatic gap finding and merging.}\label{fig:onboard_gap_finding}
	\end{subfigure}

	\begin{subfigure}{0.9\textwidth}
		\includegraphics[width = \textwidth]{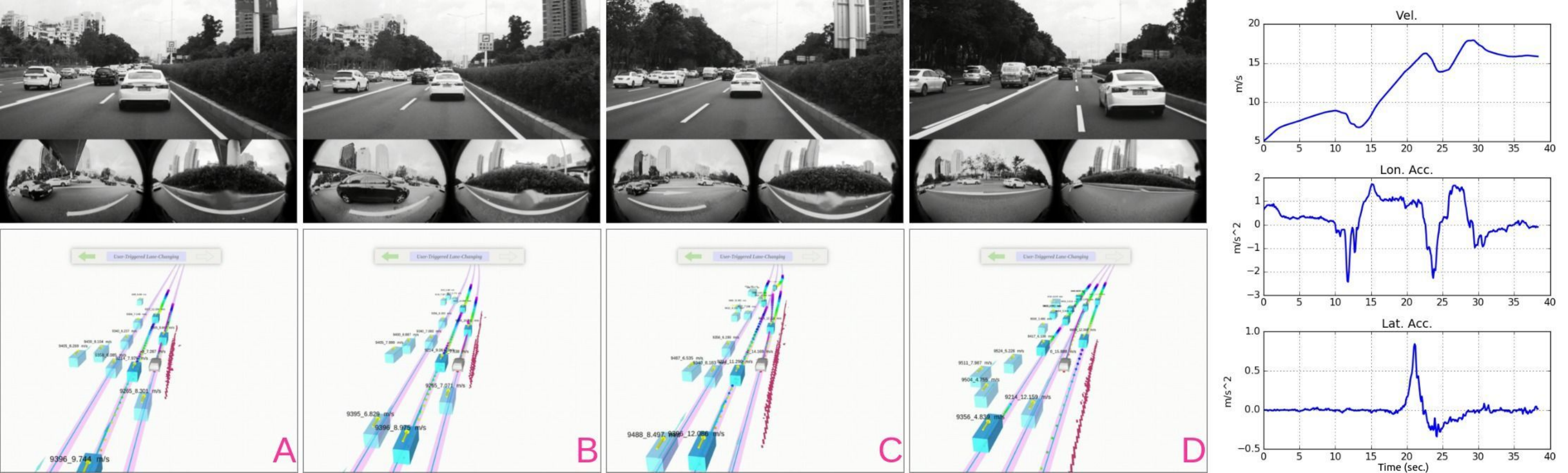}
		\caption{Interaction-aware overtaking.}\label{fig:onboard_overtaking}
	\end{subfigure}
	\caption{Illustration of two representative scenarios collected from road tests. The tracklets are represented by \textit{blue} bounding boxes, the static obstacles are marked in \textit{red} and the controlled vehicle is represented using the model vehicle. In the visualization, we illustrate the forward simulation results, both for the controlled vehicle and surrounding vehicles, using~\textit{rainbow} dots, where the color reflects the elapsed time. For a clear visualization, we only draw the forward simulation results for the most likely scenario, and the corresponding intention estimations are marked by~\textit{yellow} arrows. In these two cases, the user inputs a stick signal, which requires the planning system to conduct a lane change for the user. EPSILON automatically finds the best time and the best gap to smoothly and safely complete the merging requirement. On the right we plot the dynamic profile (velocity, longitudinal and lateral acceleration) during the process.}\label{fig:onboard_user}
\end{figure*}

In Section~\ref{sec:bp_guided_action}, we show that by incorporating advanced driving styles into semantic actions, we can fully exploit the power of semantic actions and achieve intelligent and flexible maneuvers. As shown in Fig.~\ref{fig:onboard_gap_finding}, (A) the user indicates a lane change requirement by a stick signal, and (B) the controlled vehicle moves forward and automatically finds the appropriate gap to merge. In (C), the controlled vehicle finally finds an appropriate gap and starts merging, and in (D) the controlled vehicle completes the merging. From this example, we observe that by introducing driving knowledge through intelligent driving controllers, the lane-change semantic action can be much more flexible and automatically adapts to the traffic configuration. Compared to MPDM~\cite{cunningham2015mpdm}, the maneuverability of the controlled vehicle is more flexible and intelligent, while compared to pure rule-based decision making methods, EPSILON is more general and does not rely on complex handcrafted logics. 

\subsubsection{Interaction-aware overtaking}
In Section~\ref{sec:bp_forward_integration} and Section~\ref{sec:safety_mechnism}, we present the multi-agent forward simulation and safety mechanism to consider interaction in a safe way. As illustrated in Fig.~\ref{fig:onboard_overtaking}, we show how interaction-aware overtaking can be achieved. In (A), the user indicates a lane change requirement. However, we find that the following vehicle in the neighboring lane is not cooperative. What is worse, it tries to compete with the controlled vehicle and prevent it from changing lanes (B). Actually, this is a common driving style in the area where the field tests are conducted. In (C), the ego vehicle further increases the speed and from the dynamic profile, we can see that the acceleration is even larger compared to the first phase, to increase the gap with respect to the following vehicle. This acceleration is due to the active safety mechanism conducting a longitudinal evasive maneuver so that even in the case of a bump-to-rear collision, the controlled vehicle would not be at fault since it would have already completed the lane change. The controlled vehicle is confident about conducting the acceleration maneuver since there is enough room for the backup choice to quit overtaking. In (D), the following vehicle does not further accelerate and the ego vehicle finds a safe gap to complete the merge. From this example we find that interaction-aware planning does increases the flexibility of the decision and also avoids over-conservative behaviors. However, it is important to consider interaction in the envelope of safety.

\subsubsection{Low-speed automatic lane change in dense traffic}
\begin{figure*}[t]
	\centering
	\begin{subfigure}{0.9\textwidth}
		\includegraphics[width = 0.99\textwidth]{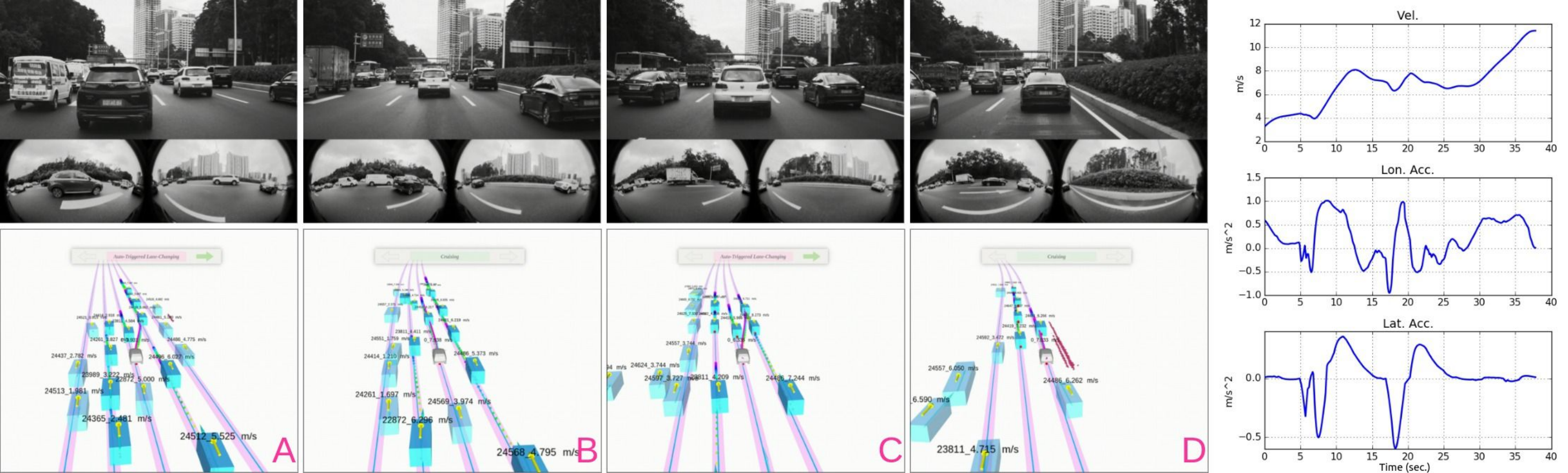}
		\caption{Low-speed automatic lane change in dense traffic.}\label{fig:onboard_alc}
	\end{subfigure}

	\begin{subfigure}{0.9\textwidth}
		\includegraphics[width = 0.99\textwidth]{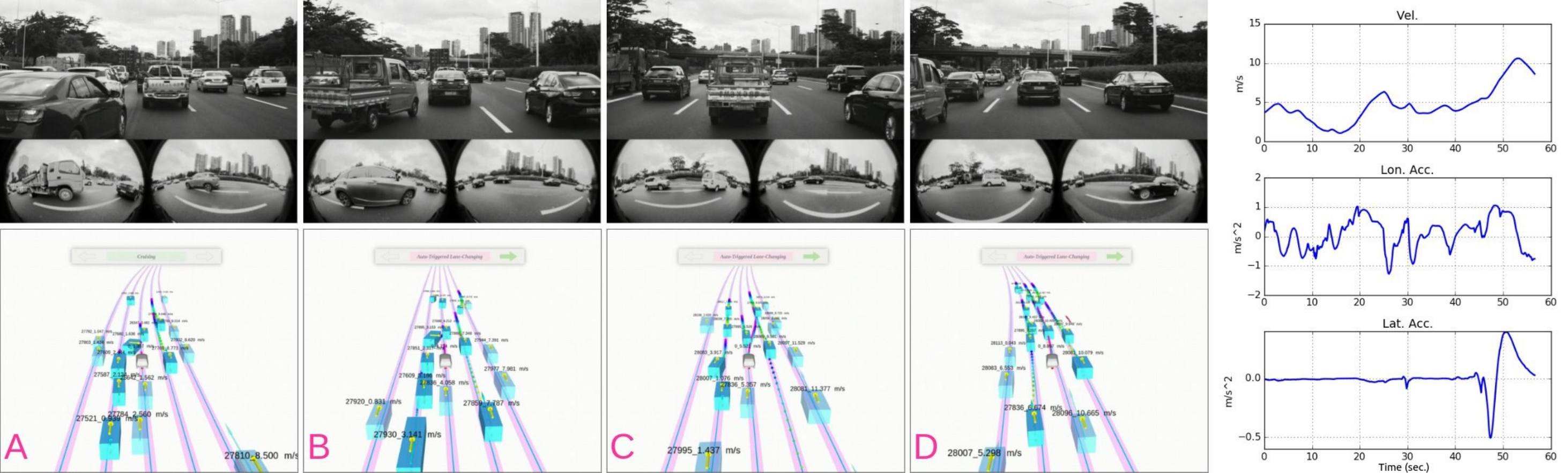}
		\caption{Cut-in handling and automatic lane change.}\label{fig:onboard_cutin_alc}
	\end{subfigure}
	\caption{Illustration of two automatic lane change scenarios in dense traffic. The visualization scheme is the same as in Fig.~\ref{fig:onboard_user}, and the difference is that the lane changes in Fig.~\ref{fig:onboard_user} are upon user's request, while here, the lane changes are proposed by the planning system.}\label{fig:onboard_automatic}
\end{figure*}

The previous two cases are lane changes upon user requests. And in this case, we show how automatic lane change can be conducted. Automatic lane change is more challenging in low-speed dense traffic.
In Fig.~\ref{fig:onboard_alc}, the controlled vehicle tries to accelerate and conduct an active merge into the nearby lane where the traveling efficiency is higher. The controlled vehicle considers the interaction with the following vehicle and finishes the lane change safely in (B). Then the controlled vehicle finds the nearby lane is more efficient than the current lane and proposes another lane change, as shown in (C). The controlled vehicle completes the second lane change safely in (D). From these two consecutive active lane changes, we find that, by considering interaction and equipping it with the safety mechanism, EPSILON can achieve quite flexible maneuvers in this dense traffic. Moreover, there are interesting points in this case concerning the observation noise. The observed velocity of the following vehicle in the nearby lane is subject to large error, especially when the controlled vehicle starts steering (see the attached video for details). Even under this observation noise, EPSILON can guarantee the safety and avoiding over-conservative behaviors.

\subsubsection{Cut-in handling and automatic lane change}
In this case we show how the controlled vehicle handles the aggressive cut-in maneuvers of other traffic participants. In Fig.~\ref{fig:onboard_cutin_alc}, the front vehicle in the nearby lane conducts an aggressive lane-change close to the controlled vehicle (A), and the controlled vehicle encountered another cut-in by a truck (B). Both cut-ins are handled smoothly and the stop-and-go process is comfortable as we can observe from the dynamic profile. Later, the controlled vehicle decides to overtake the vehicles that cut in previously since it is congested in the current lane, as shown in (C). Finally, the controlled vehicle automatically switches to a more efficient lane (D). From this case we can observe that the intention estimation and forward simulation work well in dense traffic. Cut-in maneuvers can be quickly identified and handled smoothly. Moreover, the controlled vehicle is not being too conservative and leaves space for others all the time. It also actively overtakes other vehicles for considering its own travel efficiency.

\section{Conclusion}\label{sec:conclusion}
In this paper, we present EPSILON, an efficient planning system for highly interactive environments. 
EPSILON achieves flexible maneuverability while preserving computational efficiency via guided branching, inspired by domain knowledge. A safety mechanism is introduced into multi-agent forward simulation to consider interaction in the envelope of safety. A novel motion planning technique using a spatio-temporal semantic corridor is presented to convert the decision to a safe and comfortable executable trajectory while retaining satisfactory consistency with the behavior planning. EPSILON is validated on a real vehicle in real-world dense city traffic. In real-world driving, EPSILON is neither too conservative like most traditional methods, nor over-aggressive considering noisily rational traffic participants.


\vspace{-1.0cm}

\begin{IEEEbiography}[{\includegraphics[width=1in,height=1.25in,clip,keepaspectratio]{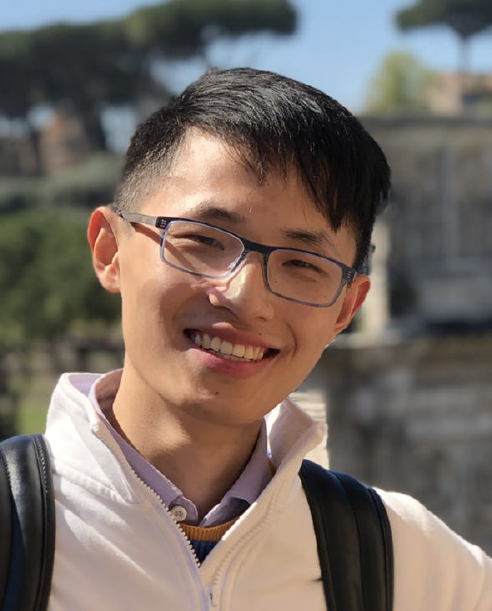}}]{Wenchao Ding}
	received his B.Eng. degree in electronic and information engineering from Huazhong University of Science and Technology, China, in 2015. He received his Ph.D. degree in electronic and computer engineering from the Hong Kong University of Science and Technology, Hong Kong, in 2020. He is currently working as a research engineer at Huawei Technology Co., Ltd.

	His research interests include decision making, prediction, motion planning and autonomous navigation for aerial robots and autonomous vehicles.
\end{IEEEbiography}

\vskip -1.8\baselineskip plus -1fil

\begin{IEEEbiography}[{\includegraphics[width=1in,height=1.25in,clip,keepaspectratio]{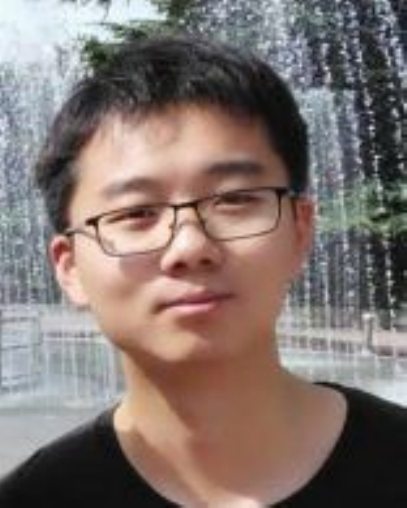}}]{Lu Zhang}
	received his B.Eng. and M.Eng. in automation from Beijing Institute of Technology, China, in 2015 and 2018. He then joined HKUST Aerial Robotics Group at the Hong Kong University of Science and Technology, under the supervision of Prof. Shaojie Shen.

	His research interests cover decision-making, motion planning, and motion prediction for autonomous vehicles.
\end{IEEEbiography}

\vskip -1.8\baselineskip plus -1fil

\begin{IEEEbiography}[{\includegraphics[width=1in,height=1.25in,clip,keepaspectratio]{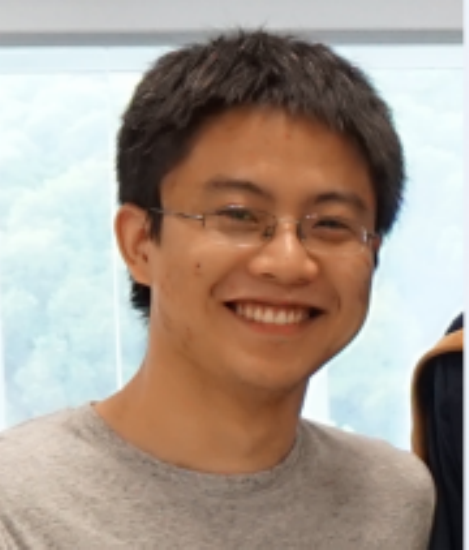}}]{Jing Chen}
	received his B.Eng. degree in computer science and technology from Harbin Institute of Technology, China, in 2014. He received his M.Phil. degree in robotics from Hong Kong University of Science and Technology, Hong Kong, in 2016. He is currently working as an algorithm engineer in DJI.

	His research interests include planning, optimization programming, mobile robot navigation.
\end{IEEEbiography}

\vskip -1.8\baselineskip plus -1fil

\begin{IEEEbiography}[{\includegraphics[width=1in,height=1.25in,clip,keepaspectratio]{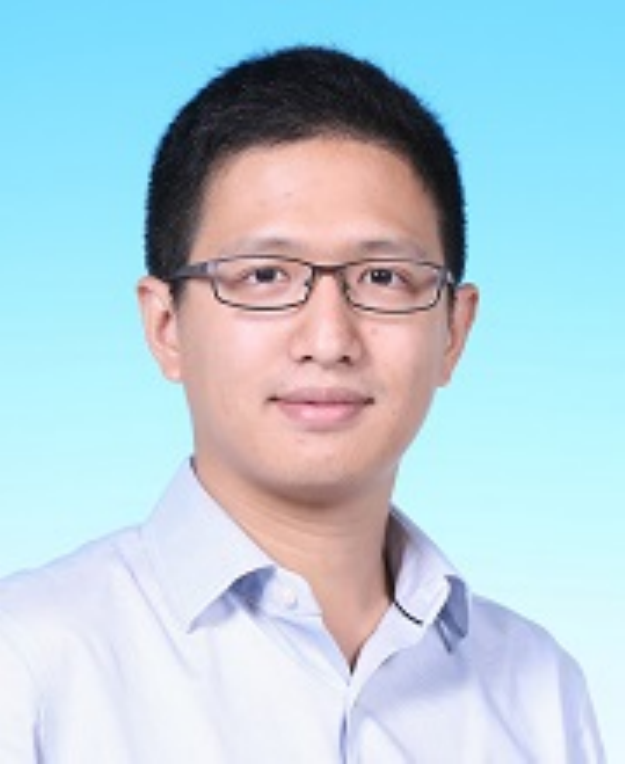}}]{Shaojie Shen}
	received his B.Eng. degree in Electronic Engineering from the Hong Kong University of Science and Technology in 2009. He received his M.S. in Robotics and Ph.D. in Electrical and Systems Engineering in 2011 and 2014, respectively, all from the University of Pennsylvania, PA, USA. He joined the Department of Electronic and Computer Engineering at the HKUST in September 2014 as an Assistant Professor, and was promoted to Associate Professor in July 2020.

	His research interests are in the areas of robotics and unmanned aerial vehicles, with focus on state estimation, sensor fusion, computer vision, localization and mapping, and autonomous navigation in complex environments.
\end{IEEEbiography}

\end{document}